\documentclass[10pt,twocolumn,letterpaper
]{article}

\usepackage[pagenumbers]{cvpr}            %

\usepackage[dvipsnames]{xcolor}

\newcommand{\xcaption}[1]{\vspace{-2mm}\caption{#1}}

\definecolor{cvprblue}{rgb}{0.21,0.49,0.74}
\usepackage[pagebackref,breaklinks,colorlinks,citecolor=cvprblue]{hyperref}

\newcommand{\paperTitle}{Sparse Views, Near Light:\\[2mm]A Practical Paradigm for Uncalibrated Point-light Photometric Stereo\vspace{-3.5mm}}

\newcommand{\paperIDCVPR}{7175}

\author{Mohammed Brahimi$^{1,2}$ \hspace{2mm}
Bjoern Haefner$^{1,2,4\footnotemark}$ \hspace{1.8mm}
Zhenzhang Ye$^{1,2}$ \hspace{2mm}
Bastian Goldluecke$^3$ \hspace{2mm}
Daniel Cremers$^{1,2}$\\ \hspace{-5mm}
$^1$ TU Munich, $^2$ Munich Center for Machine Learning, $^3$ University of Konstanz, $^4$ NVIDIA\\
{\hspace{-5mm}\tt\small$\lbrace$mohammed.brahimi, bjoern.haefner, yez, cremers$\rbrace$@tum.de}\\ {\tt\small bastian.goldluecke@uni-konstanz.de}
}

\usepackage{array} %
\usepackage{graphicx}
\usepackage[accsupp]{axessibility}
\usepackage{amsmath}
\usepackage{amssymb}
\usepackage{relsize}
\usepackage{booktabs}
\usepackage{bm}
\usepackage{blindtext}
\usepackage{balance}
\usepackage{tikz}
\usepackage{pgfplots}
\usepackage{caption}
\usepackage{subcaption}
\usepackage{stfloats}
\pgfplotsset{compat=1.16}
\usetikzlibrary{spy,arrows}
\selectcolormodel{cmyk}
\usepgfplotslibrary{colormaps}
\DeclareMathOperator*{\argmin}{arg\,min}
\newcommand{\norm}[1]{ \left\lVert#1\right\rVert }
\newcommand{\density}{\sigma}
\newcommand{\px}{\mathbf{x}}

\newcommand{\pl}{\mathbf{l}}
\newcommand{\plp}{\mathbf{p}}
\newcommand{\R}{\mathbb{R}}
\newcommand{\sdf}{d}
\newcommand{\radiance}{L}
\newcommand{\img}{I}
\newcommand{\accradiance}{\mathcal{L}}
\newcommand{\accradianceamb}{\mathcal{L}_\text{amb}}
\newcommand{\accradiancepoint}{\mathcal{L}_\text{pt}}
\newcommand{\pdf}{w}
\newcommand{\dx}[1]{\;\mathrm{d}#1}
\newcommand{\pn}{\mathbf{n}}
\newcommand{\pc}{\mathbf{c}}
\newcommand{\pp}{p}
\newcommand{\vo}{\mathbf{v}}

\newcommand{\brdf}{f_\text{r}}
\newcommand{\brdfd}{\rho}
\newcommand{\brdfs}{f_\text{s}}
\newcommand{\rough}{\alpha_\text{r}}
\newcommand{\spec}{\alpha_\text{s}}

\newcommand{\sdfparam}{\theta}
\newcommand{\brdfparam}{\gamma}
\newcommand{\diffuseparam}{\brdfparam_1}
\newcommand{\specparam}{\brdfparam_2}
\newcommand{\e}{\text{e}}

\newcommand{\dogA}{\textit{dog1}}
\newcommand{\dogB}{\textit{dog2}}
\newcommand{\girlA}{\textit{girl1}}
\newcommand{\girlB}{\textit{girl2}}
\newcommand{\bird}{\textit{bird}}
\newcommand{\squirrel}{\textit{squirrel}}
\newcommand{\hawk}{\textit{hawk}}
\newcommand{\flamingo}{\textit{flamingo}}
\newcommand{\pumpkin}{\textit{pumpkin}}
\newcommand{\rooster}{\textit{rooster}}

\newcommand{\bear}{\textit{bear}}
\newcommand{\buddha}{\textit{buddha}}
\newcommand{\pot}{\textit{pot2}}
\newcommand{\reading}{\textit{reading}}
\newcommand{\cow}{\textit{cow}}

\newcommand{\OurAlbedoNet}{\textit{OurAlbedoNet}}

\renewcommand{\S}{\mathbb{S}}

\newcommand{\subheading}[1]{{\bf #1.}}

\newcommand{\colorbar}[4][jet]{
     \begin{tikzpicture}
    \pgfmathsetmacro{\step}{#3/#4};
         \begin{axis}[
	         xmin=0,xmax=1,
	         ymin=0,ymax=1,
             hide axis,
             scale only axis,
             height=0pt,
             width=0pt,
             colormap name=#1,
             colorbar,
             point meta min=#2,
             point meta max=#3,
             colorbar style={
                 scaled y ticks = false,
                 tick label style={/pgf/number format/fixed},
                 height=6cm,
                 ytick={#2,\step,...,#3},
                 /pgf/number format/precision=3
             }]
         \end{axis}
     \end{tikzpicture}
}
\pgfplotsset{colormap={custom}{
			rgb(0cm)=(0,0,0.5)
			rgb(0.78cm)=(0,0,1.0)
			rgb(2.28cm)=(0,1.0,1.0)
			rgb(3.78cm)=(1.0,1.0,0)
			rgb(5.28cm)=(1.0,0,0)
                rgb(6cm)=(0.5,0,0)
    		}
}

\def\paperID{\paperIDCVPR} %
\def\confName{CVPR}
\def\confYear{2024}

\title{\paperTitle}

\begin{document}
\ifx\singlecolteaser\undefined
  \maketitle
\else
  \twocolumn[{%
    \renewcommand\twocolumn[1][]{#1}%
    \maketitle
    \ifx\singlecolteaser\undefined
\begin{figure}[t]
  \centering
  \small
  \newcommand{\mywidthc}{0.015\textwidth}
  \newcommand{\mywidthx}{0.115\textwidth}
  \newcommand{\mywidthi}{0.115\textwidth}
  \newcommand{\mywidthj}{0.115\textwidth}
  \newcolumntype{C}{ >{\centering\arraybackslash} m{\mywidthc} }
  \newcolumntype{X}{ >{\centering\arraybackslash} m{\mywidthx} }
  \newcommand{\tabelt}[1]{\hfil\hbox to 0pt{\hss #1 \hss}\hfil}
  \setlength\tabcolsep{1pt} %
  \begin{tabular}{XXXX}
    \includegraphics[width=\mywidthi]{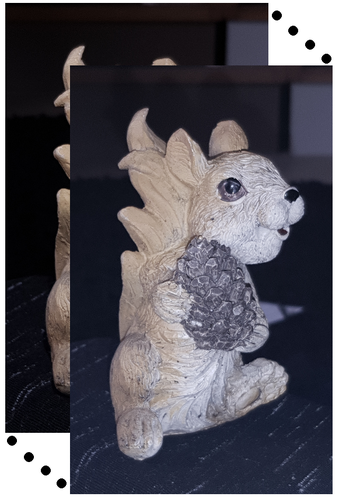}&
    \includegraphics[width=\mywidthi]{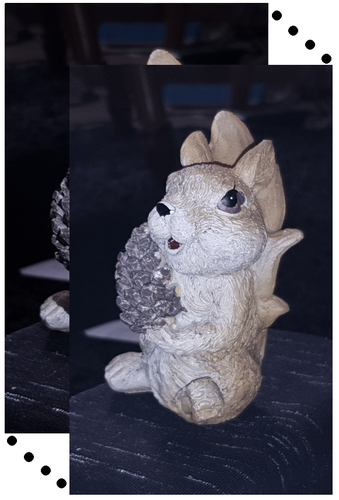}&
    \includegraphics[width=\mywidthj]{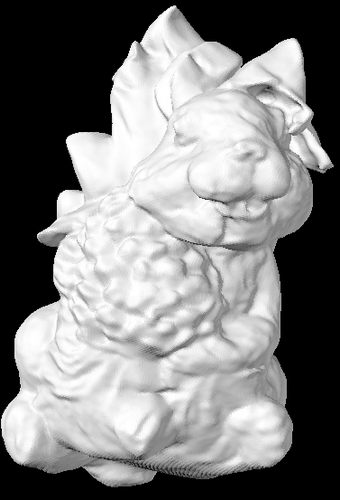}
 &
    \includegraphics[width=\mywidthj]{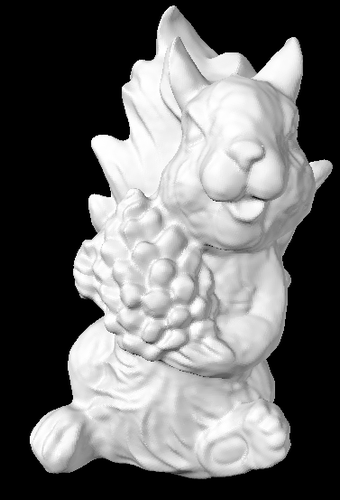}   \\
    Multi-illumination data at viewpoint 1 & Multi-illumination data at viewpoint 2&PS-NeRF~\cite{yang2022ps}&Ours
  \end{tabular}
  \vspace{-3mm}
  \caption{\small
  We introduce the first framework for multi-view uncalibrated point-light photometric stereo. Given a set of PS images captured from different viewpoints (left), our method recovers high-fidelity $3$D reconstruction (right).
  The acquisition of uncalibrated point-light PS imagery captured under ambient lighting in a sparse multi-view setup does not only allow for easy data capture, but also leads to $3$D reconstructions of unprecedented detail.
  Here, with as few as two views we are able to reconstruct the squirrel's $3$D geometry at higher precision than the state-of-the-art.
  }
  \vspace{-1mm}
\label{fig:teaser}
\end{figure}

\else

\begin{center}
  \captionsetup{type=figure}
  \small
  \newcommand{\mywidthc}{0.02\textwidth}
  \newcommand{\mywidthx}{0.15\textwidth}
  \newcolumntype{C}{ >{\centering\arraybackslash} m{\mywidthc} }
  \newcolumntype{X}{ >{\centering\arraybackslash} m{\mywidthx} }
  \newcommand{\tabelt}[1]{\hfil\hbox to 0pt{\hss #1 \hss}\hfil}
  \setlength\tabcolsep{1pt} %
  \begin{tabular}{CXXCXX}
    \rotatebox{90}{Input}&
    \includegraphics[width=\mywidthx]{images/teaser_images/viewpoint_1.png}&
    \includegraphics[width=\mywidthx]{images/teaser_images/viewpoint_2.png}&
    \rotatebox{90}{Output}&
    \includegraphics[width=\mywidthx]{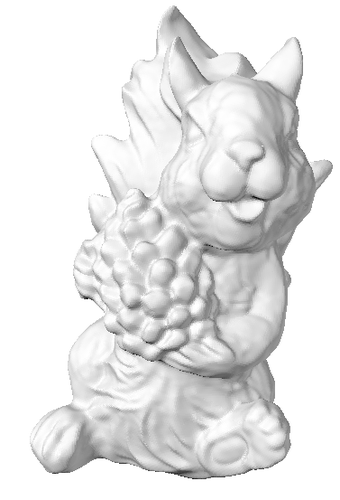}&
    \includegraphics[width=\mywidthx]{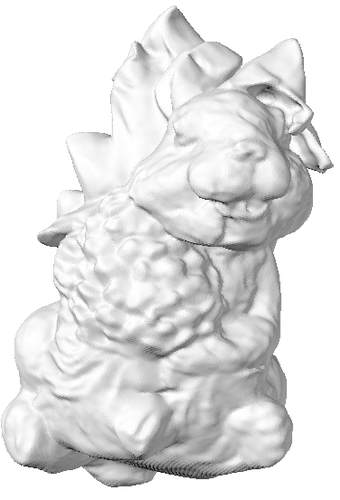}\\
    &Viewpoint 1 & Viewpoint 2&& Ours &\cite{yang2022ps} 
  \end{tabular}
  \captionof{figure}{
  Given a set of photometric stereo images from a set of different viewpoints, our proposed end-to-end framework recovers a consistent $3$D geometry of unprecedented detail.
  The combination of uncalibrated point-light photometric stereo imagery captured under ambient lighting in a sparse multi-view setup does not only allow for easy data capture, but also results in high-fidelity $3$D reconstructions.
  Here, with as few as two views we are able to reconstruct the squirrel's $3$D geometry of higher quality and detail compared to the state-of-the-art.
  }
\label{fig:teaser}
\end{center}
\fi
  }]
\fi
\footnotetext{\textsuperscript{*}The contribution was done while at TUM.}
\begin{abstract}
\vspace{-3mm}
Neural approaches have shown a significant progress on camera-based reconstruction. 
But they require either a fairly dense sampling of the viewing sphere, or pre-training on an existing dataset, thereby limiting their generalizability.
In contrast, photometric stereo (PS) approaches have shown great potential for achieving high-quality reconstruction under sparse viewpoints. Yet, they are impractical because they typically require tedious laboratory conditions, are restricted to dark rooms, and often multi-staged, making them subject to accumulated errors.
To address these shortcomings, we propose an end-to-end uncalibrated multi-view PS framework for reconstructing high-resolution shapes acquired from sparse viewpoints in a real-world environment.  We relax the dark room assumption, and allow a combination of static ambient lighting and dynamic near LED lighting, thereby enabling easy data capture outside the lab. Experimental validation confirms that it outperforms existing baseline approaches in the regime of sparse viewpoints by a large margin. This allows to bring high-accuracy 3D reconstruction from the dark room to the real world, while maintaining a reasonable data capture complexity.
\end{abstract}   

\vspace{-5mm}
\section{Introduction}
\label{sec:intro}
\ifx\singlecolteaser\undefined
  
\fi
\vspace{-1mm}
The challenge of $3$D reconstruction stands as a cornerstone in both computer vision and computer graphics.
Despite notable progress in recovering an object's shape from dense image viewpoints, predicting consistent geometry from sparse viewpoints remains a difficult task.
Contemporary approaches employing neural data structures depend heavily on extensive training data to achieve generalization in the context of sparse views.
Additionally, the presence of a wide baseline or textureless objects form significant obstacles.
In contrast, photometric methodologies like photometric stereo (PS) excel in reconstructing geometry, even in textureless regions.
This capability is attributed to the abundance of shading information derived from images acquired under diverse illumination.
Yet, such approaches typically require a controlled laboratory setup to fulfill the necessary dark room and directional light assumptions.
As a consequence, PS approaches become impractical beyond the confines of a laboratory.

To address these shortcomings we combine state-of-the-art volume rendering formulations with a sparse multi-view photometric stereo model. In particular, we advocate a physically realistic lighting model that combines ambient light and uncalibrated point-light illumination.
Our approach facilitates simplified data acquisition, and we introduce a novel pipeline capable of reconstructing an object's geometry from a sparse set of viewpoints, even if the object is completely textureless.
Furthermore, since we assume a point-light model instead of distant directional lighting, we can infer absolute depth from a single viewpoint, allowing us to address the challenge of wide camera baselines.

\noindent In detail, our contributions are as follows:
\begin{itemize}
    \item We introduce the first framework for multi-view uncalibrated point-light photometric stereo. It combines a state-of-the-art volume rendering formulation with a physically realistic model of ambient light and point lights.
    \item We eliminate the need for a dark room environment, dense capturing process, and distant lighting. Thereby we enhance the accessibility and simplify data acquisition for setups beyond traditional laboratory settings.
    \item We validate that it can be successfully used for accurate shape reconstruction of textureless objects in highly sparse scenarios with wide baselines.
    \item Despite the absence of pre-processing or vast training data, we outperform cutting-edge
    approaches that either rely only on static ambient illumination or PS imagery.
\end{itemize}
\vspace{-2mm}
\section{Related Work}
\label{sec:Related_work}
\vspace{-1mm}
In this section, we categorize pertinent works into two distinct groups.
The first category focuses on neural $3$D reconstruction, emphasizing the use of images captured under static ambient illumination.
Subsequently, the second category delves into photometric stereo (PS) approaches, which specifically involve methods assuming images captured under varying illumination conditions.
\vspace{-1mm}
\subsection{Neural 3D reconstruction}
\vspace{-2mm}
In our context, the domain of neural 3D reconstruction under static ambient light can be bifurcated into two distinct subcategories.
The first comprises methodologies that presume a dense sampling of the viewing sphere~\cite{niemeyer2020differentiable, yariv2020multiview, yariv2021volume, wang2021neus, mildenhall2020nerf, oechsle2021unisurf, fu2022geo, zhang2021learning, darmon2022improving, brahimi2022supervol, cheng2023wildlight}.
Conversely, the second subcategory encompasses techniques adept at handling sparse camera viewpoints~\cite{yao2018mvsnet, cheng2020deep, gu2020cascade, yang2020cost, zhang2020visibility, ding2022transmvsnet, wei2021aa, yan2020dense, yao2019recurrent, ren2023volrecon, liang2023retr, xu2023c2f2neus, long2022sparseneus, peng2023gens, huang2023sc, Wu_2023_ICCV, yu2022monosdf, vora2023divinet, cerkezi2023sparse}.

Given our emphasis on a sparse set of camera viewpoints, our focus will be directed toward the latter set of methodologies.
The majority of sparse reconstruction approaches within this category hinge on training with a specific dataset~\cite{yao2018mvsnet, cheng2020deep, gu2020cascade, yang2020cost, zhang2020visibility, ding2022transmvsnet, wei2021aa, yan2020dense, yao2019recurrent, ren2023volrecon, liang2023retr, xu2023c2f2neus}.
The reliance on extensive training data renders these techniques susceptible to errors when confronted with objects not represented in the training dataset, thereby introducing sensitivity to the domain gap between training and test data.

Efforts have been undertaken to mitigate the challenges associated with the domain gap.
One approach involves fine-tuning pre-trained architectures during test time with sparse input data~\cite{long2022sparseneus, peng2023gens, huang2023sc}.
However, while fine-tuning enables fine-scale refinement of the predicted reconstruction, the broader issue of generalization to entirely different datasets persists as a challenge.
Furthermore, it is crucial to note that all the previously mentioned sparse approaches presume small camera baselines and textured objects.
To achieve a full reconstruction of an object from sparse views, it becomes imperative to develop texture-agnostic approaches capable of handling %
wide baselines.

Another set of approaches involves the utilization of pre-training to establish a geometric prior~\cite{Wu_2023_ICCV, yu2022monosdf, vora2023divinet}.
Yu \etal~\cite{yu2022monosdf} leverages this geometric prior to predict a depth and normal map, albeit with a limitation in fine-scale details due to its coarse resolution.
Notably,~\cite{Wu_2023_ICCV, vora2023divinet} address this limitation.
Wu \etal~\cite{Wu_2023_ICCV} relax the assumption on small baselines, although a reasonable overlap of sparse views remains necessary for consistent 3D reconstruction.
Vora \etal~\cite{vora2023divinet} demonstrate promising results with much less overlap, employing only three viewpoints.
However, the use of template shapes as a prior term introduces the risk of the reconstruction quality being affected by the domain gap curse.

In contrast, our approach eliminates the requirement for training data and operates effectively in wide baseline scenarios, independent of the object's texture.
This accomplishment can be attributed to the integration of PS principles, which we will elucidate in the subsequent discussion.
\vspace{-3mm}
\subsection{Photometric Stereo}
\vspace{-1mm}
The principle of PS~\cite{woodham1980photometric} has a long and established history, renowned for yielding high-quality and finely detailed geometric estimates, due to its capturing process: multiple images under different illumination of a static object are captured from a single camera viewpoint.

In our case, PS approaches can thus be categorized into two main types.
The first category pertains to single-view methods~\cite{shi2016benchmark, ju2022deep, papadhimitri2014uncalibrated, queau2017semi, logothetis2017semi, mecca2014near, queau2018led, logothetis2016near, logothetis2023cnn, guo2022edge}, wherein data is acquired as described above.
The second category involves multi-view methods~\cite{li2020multi, zhao2023mvpsnet, asthana2022neural, cheng2022diffeomorphic, logothetis2019differential, hernandez2008multiview, wu2010fusing, park2016robust, kaya2023multi, kaya2022uncertainty, yang2022ps}, where the capturing process is repeated at multiple camera viewpoints.

Within the context of our task, we concentrate on the multi-view scenario.
Many existing approaches in this domain assume knowledge of the light source for each image, earning them the designation of calibrated PS (CPS).
In this case, frequently the light is parameterized as directional~\cite{li2020multi, zhao2023mvpsnet, asthana2022neural, cheng2022diffeomorphic}, while approaches employing point-light illumination are more sparsely represented~\cite{li2020multi, logothetis2019differential, cheng2022diffeomorphic}, owing to the intricate model and the much more complex calibration process~\cite{queau2018led}.
In a broader context, the calibration of each image's light source is impractical and tedious, underscoring the advantage of the uncalibrated case, which inherently facilitates a more straightforward and streamlined capturing process.

Efforts have been dedicated to exploring the uncalibrated PS (UPS) scenario, where there is no prior knowledge of the light source~\cite{hernandez2008multiview, wu2010fusing, park2016robust, kaya2023multi, kaya2022uncertainty, yang2022ps}.
Nevertheless, all these uncalibrated methods typically assume a distant and directional light model or permit a lighting model based on spherical harmonics~\cite{basri2003lambertian,Brahimi2020,Peng_2017_ICCV,haefner2019iccv,8738841,10030841,9093491}, albeit at the expense of being restricted to Lambertian objects~\cite{wu2010fusing}.
Consequently, the exploration of the uncalibrated point-light case remains uncharted territory, particularly in the multi-view scenario.

Despite the impractical assumption of calibrated light or the constraint of uncalibrated directional lighting, the aforementioned multi-view photometric stereo approaches demand a dark room and lack accommodation for ambient light~\cite{li2020multi, zhao2023mvpsnet, asthana2022neural, cheng2022diffeomorphic, logothetis2019differential, hernandez2008multiview, park2016robust, kaya2023multi, kaya2022uncertainty, yang2022ps}.
Additionally, they comprise multiple stages, rendering them prone to initialization and accumulation errors~\cite{li2020multi, logothetis2019differential, hernandez2008multiview, wu2010fusing, park2016robust, kaya2023multi, kaya2022uncertainty, yang2022ps}.
Furthermore, these methods assume Lambertian objects~\cite{logothetis2019differential, wu2010fusing} and crucially rely on a dense sampling of camera viewpoints around the object~\cite{li2020multi, logothetis2019differential, hernandez2008multiview, wu2010fusing, park2016robust}.

The approach most closely related to ours is the sparse method PS-NeRF~\cite{yang2022ps}, which assumes uncalibrated directional lighting with imagery captured in a dark room.
This method comprises multiple stages, where initially, a pre-trained network~\cite{chen2019self} is utilized to estimate the initial light, along with normal information for each image and camera view. Subsequently, this acquired information is employed to estimate the object's mesh through the optimization of an occupancy field~\cite{oechsle2021unisurf}.
The outputs of both stages are then combined to derive a consistent normal map of the object.
Although PS-NeRF demonstrates impressive results in a sparse scenario with five viewpoints, its susceptibility to error accumulation is a potential limitation, given the involvement of multiple stages.

Our proposed model addresses the aforementioned limitations comprehensively.
Notably, we excel in handling sparse yet wide baseline scenarios, where existing state-of-the-art methods encounter challenges.
Our investigation tackles the open challenge of multi-view uncalibrated point-light PS within an end-to-end framework, liberating us from the constraints associated with Lambertian objects.
Furthermore, our approach facilitates easy data capture in the presence of ambient light, eliminating the need for a dark room.
\vspace{-5mm}
\section{Setting and image formation model}
\label{sec:Method}
We consider an object under static illumination photographed from several different viewpoints.
For every viewpoint, we are given a set of $N+1$ images $(I^l)_{l=0,\dots,N}$. Image $I^0$ is captured only under static ambient illumination and called the ambient image.
For images $I^1$ to $I^N$, the object is in addition to the ambient illumination also illuminated by
a single achromatic point-light source with scalar light intensity $L_0\in\R^+$, placed at a different location~$\plp^l\in\R^3$ 
for each image and viewpoint.
We use the common assumption in volume rendering \cite{yariv2021volume,Wu_2023_ICCV,mildenhall2020nerf,wang2021neus} that the image brightness~$I^l$ is the same as the accumulated radiance of the volume rendering, see \cref{eq:volume_integral}.
Note that we do not add an index for the viewpoint to the image to not clutter notation, whenever we later evaluate an image at a pixel location, we assume it implicitly means a pixel at a specific viewpoint. In particular, a sum over all pixels is the sum over all pixels at every viewpoint location.

According to the linearity of the radiance, the total radiance for the image $l$ is given by a sum
$\accradiance^l = \accradianceamb + \accradiancepoint^{l}$,
where $\accradianceamb$ is the radiance due to the static illumination, the same for each image,
and $\accradiancepoint^{l}$ is the radiance due to the point-light source used for that particular image.
We will first express the radiance due to the point-light source with respect to the material and lighting parameters,
which allows to take into account the different illuminations of the images. Subsequently, we elaborate on how
we deal with the ambient component.
\subsection{Point-light illumination and material}
\label{sec:method_image_formation}
The %
field~$\radiance(\px,\pn,\vo)$ 
maps 
points~$\px\in\R^3$, normal direction~$\pn(\px)\in\S^2$ and viewing direction~$\vo\in\S^2$ to its radiance.
It can be expressed in terms of geometry, material and lighting with the classical rendering
equation~\cite{kajiya1986rendering}.
Since for now we only consider a single achromatic point-light source with intensity~$L_0$
at position~$\plp$, this equation simplifies to
\begin{equation}
  \radiance(\px,\pn,\vo) = \dfrac{L_0}{\norm{\plp - \px}^2} \brdf(\px,\pn,\vo,\pl) \max(0,\pn \cdot \pl),
  \label{eq:point_light_model}
\end{equation}
where $\pl = \frac{\plp - \px}{\norm{\plp - \px}}$.

As a model for the spatially varying %
reflectance~$\brdf$, %
we use the simplified Disney BRDF \cite{karis2013real}.
It is relatively simple yet expressive enough to resemble a wide variety of materials.
It has already been successfully used in several other prior
works~\cite{luan2021unified, zhang2021physg, brahimi2022supervol}.
The Disney BRDF requires three parameters to be available at every point %
$\px\in\R^3$:
An RGB diffuse albedo $\rho \in \R^3_{\geq0}$, a roughness~$\rough>0$ and a specular
albedo~$\spec\in [0,1]$.

As in~\cite{brahimi2022supervol}, all three BRDF parameters can be represented with MLPs.
One MLP with network parameters $\diffuseparam$ computes the diffuse parameter~$\rho$,
a second MLP with 
network parameters $\specparam$ computes the specular parameters~$(\rough,\spec)$.
To encompass all point-light and BRDF parameters, we concatenate them into the vectors $\phi = (\plp^1, \dots, \plp^M)$, which includes all light source positions (where $M$ is the product of $N$ with the number of viewpoints), and $\brdfparam = (\diffuseparam, \specparam)$, representing the BRDF parameters.
Both $\phi$ and $\brdfparam$ serve as the network parameters to be learned during optimization.
However, note that one of our %
contributions is a novel strategy to deal with the diffuse albedo
later on in \cref{sec:opt_diffuse_albedo},
which dramatically improves performance in extremely sparse scenarios.

We assume that the intensity $L_0$ remains constant across \textit{all} images and views.
This assumption is reasonable, \eg, when using a smartphone's flashlight or a sole LED, as shown in our experiments.
A constant intensity introduces a scale ambiguity between the $\brdf$ and $L_0$.
We set $L_0 \equiv 1$ and subsequently estimate a scaled version of the BRDF, \ie, the intensity is absorbed into the BRDF.
This is feasible, since we can scale the Disney BRDF arbitrarily. %

\subsection{Ambient light}
Given the provided ambient image, we adopt the approach from~\cite{zhang2012edge}, which involves subtracting the ambient image from all other images.
This process effectively eliminates ambient radiance from our considerations, simulating a traditional dark room scenario.
A drawback is that in practice, the captured images contain additive noise,
hence such a subtraction will double the variance of the noise distribution,
resulting in noisier input images.
Nevertheless, as can be seen in the results, our approach works successfully with this strategy even if we use %
a simple smartphone camera for the data capture.

However, this simplification is enabled by the fact that the point-light sources are near the object, making it easy to obtain a well exposed contribution from the point-light source.
In methods assuming directional lighting, the practical simulation often involves using a point-light source placed far from the object, ensuring that the object's diameter is significantly smaller than the distance between the point-light and the object.
For example, for the DiLiGenT-MV dataset \cite{li2020multi}, the light sources are roughly 1m from the object, whereas for our setup, the distance is approximately 25cm. As a consequence, for a given static ambient illumination, the distant point-light will need to be $16\times$ brighter in the Diligent setup to obtain a similar signal to noise ratio. %
This increased brightness requirement escalates the hardware demands, especially if the highest quality is sought.

\subsection{Other illumination effects}
We do not model indirect lighting and %
cast-shadows explicitly. However, multiple previous works~\cite{bi2020deep, haefner2021recovering, luan2021unified, nam2018practical, iron-2022}
and PS frameworks~\cite{queau2018led, guo2022edge, li2020multi, cheng2022diffeomorphic} have
demonstrated that satisfactory results can be achieved without taking these into account.
In addition, we use the robust $L^1$-norm in our data term~\eqref{eq:data_term} to further increase robustness.
Furthermore, we will expose a simple strategy in \cref{sec:optimization} to reduce the impact of %
cast-shadows, which inevitably occur when capturing a nonconvex object under changing illumination.

\section{Volume rendering of the surface}
\label{sec:Preliminaries}
For differentiable rendering we build upon VolSDF~\cite{yariv2021volume} because in contrast to NeRF~\cite{mildenhall2020nerf} it allows us to decouple geometric
representation and appearance.
As discussed in~\cite{Wu_2023_ICCV,liang2023retr,ren2023volrecon}, VolSDF exhibits poor reconstruction quality in the sparse viewpoint scenario due to the
shape-radiance ambiguity.
We eliminate this ambiguity thanks to photometric stereo, providing shading cues from different
illuminations for each viewpoint.
To this end, we employ the radiance field in \cref{eq:point_light_model}
and apply it to the multi-view uncalibrated point-light PS setting.
The integration of differentiable volume rendering with a physically realistic lighting model provides an efficient solution to the multi-view photometric stereo problem, enabling high-quality reconstruction even with sparse viewpoints, as demonstrated in the subsequent results.

\subheading{Geometry model and associated density}
The geometry of the scene is modeled with a signed distance function (SDF)
$\sdf:\Omega\to\R$ within the volume $\Omega\subset\R^3$.
We follow~\cite{park2019deepsdf,yariv2020multiview,wang2021neus,yariv2021volume} and represent the SDF with an MLP $\sdf(\px;\sdfparam)$ parametrized by~$\sdfparam$.
Note that we assume positive values of the SDF~$d$ inside the object, thus the
normal vector is given by~$\pn=\nabla \sdf/\norm{\nabla \sdf}$ on $\Omega$.
To render the scene, we follow~\cite{yariv2021volume} and 
transform the SDF into a density~$\density:\Omega\to\R_{\geq0}$
by means of the transformation
\begin{equation}
\label{eq:preliminaries:density}
\density(\px) = \beta^{-1}\Psi_\beta(-\sdf(\px)),
\end{equation}
where $\Psi_\beta$ denotes the cumulative distribution function of the Laplace distribution with zero mean. The positive scale parameter~$\beta$ is learned during optimization. For simplicity, %
we assume it is included in the set of SDF parameters~$\sdfparam$.

\subheading{Volume Rendering}
Using the density~\eqref{eq:preliminaries:density}, we can now
set up a distribution of weights $\pdf(t)$
along each ray~$\px(t) = \pc + t \vo, t \geq 0$ defined by the center of projection~$\pc$
of the camera and viewing direction~$\vo$,
\begin{equation}
  \pdf(t)=\density(\px(t))\,\exp\left( -\int_0^t\density(\px(s))\dx{s} \right ).
\label{eq:weights}
\end{equation}
This is the derivative of the opacity function, which is monotonically increasing
from zero to one, and thus~$\pdf$ is indeed a probability distribution~\cite{yariv2021volume}.
In particular, the integral
\begin{equation}
  \accradiance_\pp = \int_{0}^\infty \pdf(t) \radiance(\px(t),\pn(t),\vo)\dx{t}
  \label{eq:volume_integral},
\end{equation}
to compute radiance~$\accradiance_\pp$ in the pixel~$\pp\in\R^2$ corresponding to the ray $\px(t)$
is well-defined. Intuitively, we accumulate the radiance field~\eqref{eq:point_light_model} of the point-light source
at locations where opacity increases, i.e.
where we move towards the inside of the object. For an ideal sharp boundary, $w$ would be a
Dirac distribution with its peak placed at the boundary.

As was done in~\cite{yariv2021volume} we approximate the integral
\eqref{eq:volume_integral}
using the quadrature rule
\begin{equation}
  \accradiance_p \approx \sum^{m-1}_{i=1} (t_{i+1}-t_i)\pdf(t_i)L(\px(t_i),\pn(t_i),\vo).
  \label{eq:discrete_volume_integral}
\end{equation}
for numerical integration with a discrete set of samples~$t_1 < t_2 < \dots < t_{m}$.
Note that the integral~\eqref{eq:weights} required for~$\pdf(t_i)$
in~\eqref{eq:discrete_volume_integral}
can be numerically computed using an analogous formula
while iteratively accumulating the sum~\eqref{eq:discrete_volume_integral}.

\section{Training objective and optimization}
\label{sec:optimization}
In order to reconstruct the scene by inverse rendering, we train our neural networks from the available
input images. 
We optimize for the geometry parameters $\sdfparam$ of the SDF network
and point-light and BRDF parameters $\phi$ and $\brdfparam$ of the appearance network, respectively.

\subsection{Vanilla loss function with MLP for albedo}
\label{sec:vanilla}
The overall loss consists of the sum of an inverse rendering loss~$E_\text{RGB}(\sdfparam, \phi, \brdfparam)$,
an Eikonal loss~$E_\text{eik}(\sdfparam)$ and a mask loss~$E_\text{mask}(\sdfparam)$ for the SDF,
\begin{equation}
  E(\sdfparam,\phi,\brdfparam) = E_\text{RGB}(\sdfparam,\phi,\brdfparam) + \lambda_1 E_\text{eik}(\sdfparam) + \lambda_2 E_\text{mask}(\sdfparam),
  \label{eq:loss_function}
\end{equation}
with respective weighting parameters $\lambda_1, \lambda_2\geq0$ given as hyperparameters.
We will describe the different terms in detail in the following.

\begin{figure}[t]
  \centering
  \small
  \newcommand{\mywidthc}{0.02\textwidth}
  \newcommand{\mywidthx}{0.155\textwidth}
  \newcolumntype{C}{ >{\centering\arraybackslash} m{\mywidthc} }
  \newcolumntype{X}{ >{\centering\arraybackslash} m{\mywidthx} }
  \newcommand{\tabelt}[1]{\hfil\hbox to 0pt{\hss #1 \hss}\hfil}
  \setlength\tabcolsep{1pt} %
  \begin{tabular}{XXX}
  \begin{tikzpicture}[spy using outlines={rectangle,connect spies}]
      \node[anchor=south west,inner sep=0]  at (0,0) {\includegraphics[width=\mywidthx]{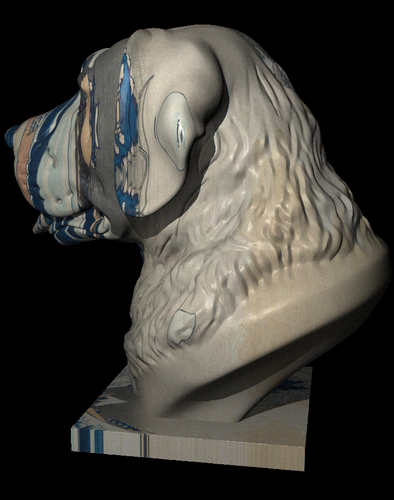}};
      \spy[color=green,width=1.2cm,height=1.2cm,magnification=2.2] on (1.05,2.0) in node [right] at (0.01,0.61);
    \end{tikzpicture} &
    \begin{tikzpicture}[spy using outlines={rectangle,connect spies}]
      \node[anchor=south west,inner sep=0]  at (0,0) {\includegraphics[width=\mywidthx]{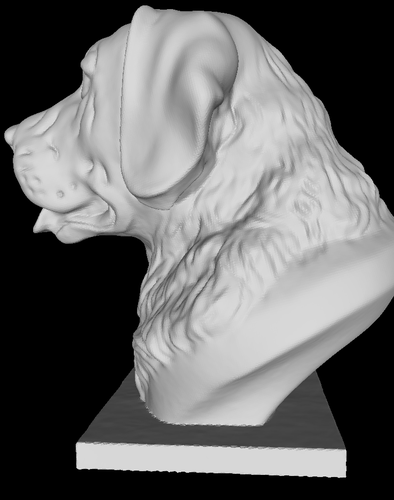}};
      \spy[color=green,width=1.2cm,height=1.2cm,magnification=2.2] on (1.1,2.15) in node [right] at (0.01,0.61);
    \end{tikzpicture} & \begin{tikzpicture}[spy using outlines={rectangle,connect spies}]
      \node[anchor=south west,inner sep=0]  at (0,0) {\includegraphics[width=\mywidthx]{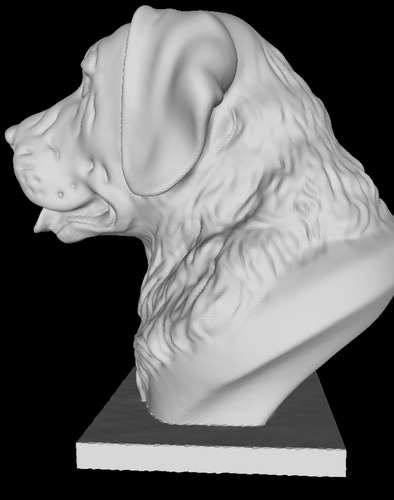}};
      \spy[color=green,width=1.2cm,height=1.2cm,magnification=2.2] on (1.1,2.15) in node [right] at (0.01,0.61);
    \end{tikzpicture}\\
    Input image & w/o truncation & w/ truncation
  \end{tabular}
  \xcaption{Effect of cast-shadows on the result. The region behind the ear exhibits frequent cast-shadows in the input images (left), which induce artefacts in the estimated mesh (middle). Our truncation strategy (\cref{sec:vanilla}) successfully eliminates these (right). }
\label{fig:self-shadows}
\end{figure}

The \textbf{rendering loss}
\begin{equation}
  E_\text{RGB}(\sdfparam,\phi,\brdfparam) = \sum_\pp \sum_{l \in B_\tau(\pp)} \norm{\img^l_\pp -\accradiance^l_\pp(\sdfparam,\phi,\brdfparam)}_1 
  \label{eq:data_term}
\end{equation}
encourages that the rendered images resemble the input images.
As a strategy to deal with cast-shadows, %
for each pixel~$\pp$, we use only the $\tau$ brightest views
$B_\tau(\pp) \subset \lbrace 1\hdots N \rbrace$, where
$\tau>0$ is a hyper-parameter.
This reduces the impact of cast-shadows, %
as a pixel will have the smallest possible
brightness when it lies in shadow.
Such a strategy is necessary since despite the use of an $L^1$-norm to improve
robustness against outliers, some regions with frequent cast-shadows %
would still exhibit small artefacts, as can be seen in \cref{fig:self-shadows}.
In order to avoid wasting meaningful information due to this heuristic,
we set $\tau=N$ for the first half of
the iterations, and then set it to $\tau=\lceil \frac{3}{4}N \rceil$ for the second half for
improved handling of cast-shadows. %

The \textbf{Eikonal term}
\begin{equation}
  E_\text{eik}(\sdfparam) = \sum_\px (\norm{\nabla_\px \sdf(\px;\sdfparam)} - 1 )^2
  \label{eq:eikonal_term}
\end{equation}
encourages $\sdf(\px;\sdfparam)$ to approximate an SDF, as it penalizes deviations from
the respective partial differential equation which every SDF satisfies,
see e.g.~\cite{gropp2020implicit}.

Finally, we follow~\cite{brahimi2022supervol,wang2021neus}
and employ the \textbf{mask loss}
\begin{equation}
  E_\text{mask}(\sdfparam) = \sum_\pp \text{BCE}(M_\pp,W_\pp(\sdfparam)),
  \label{eq:mask_term}
\end{equation}
to impose silhouette consistency by means of the binary cross entropy (BCE)
between the given binary mask $M_\pp$ at the pixel $\pp$ and
the sum $W_\pp(\sdfparam) = \sum_{i=1}^{m-1} \pdf(t_i;\sdfparam)$
of the weights at the sampling locations~$t_i$ used in~\eqref{eq:discrete_volume_integral}.

\begin{figure*}[t]
  \small
  \newcommand{\mywidthc}{0.02\textwidth}
  \newcommand{\mywidthx}{0.164\textwidth}
  \newcolumntype{C}{ >{\centering\arraybackslash} m{\mywidthc} }
  \newcolumntype{X}{ >{\centering\arraybackslash} m{\mywidthx} }
  \newcommand{\tabelt}[1]{\hfil\hbox to 0pt{\hss #1 \hss}\hfil}
  \setlength\tabcolsep{1pt} %
  \begin{tabular}{XXXXXX}
    \begin{tikzpicture}[spy using outlines={rectangle,connect spies}]
      \node[anchor=south west,inner sep=0]  at (0,0) {\includegraphics[width=\mywidthx]{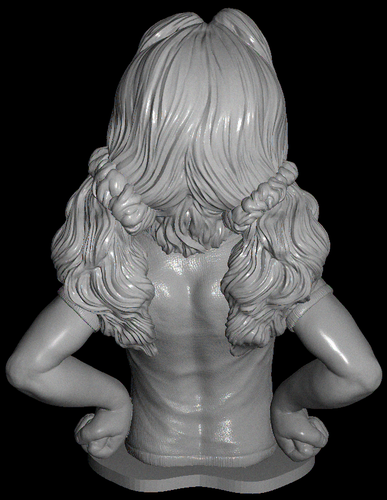}};
      \spy[color=green,width=0.7cm,height=0.9cm,magnification=1.3] on (1.4,2.15) in node [right] at (0.05,3.15);
    \end{tikzpicture}& 
    \begin{tikzpicture}[spy using outlines={rectangle,connect spies}]
      \node[anchor=south west,inner sep=0]  at (0,0) {\includegraphics[width=\mywidthx]{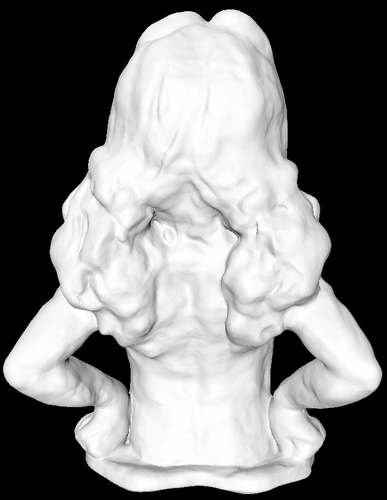}};
      \spy[color=green,width=0.7cm,height=0.9cm,magnification=1.3] on (1.4,2.15) in node [right] at (0.05,3.15);
    \end{tikzpicture}&  
    \begin{tikzpicture}[spy using outlines={rectangle,connect spies}]
      \node[anchor=south west,inner sep=0]  at (0,0) {\includegraphics[width=\mywidthx]{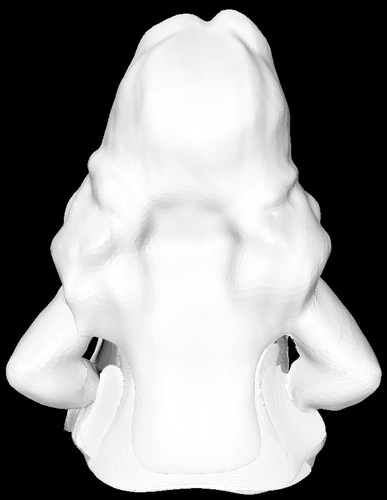}};
      \spy[color=green,width=0.7cm,height=0.9cm,magnification=1.3] on (1.4,2.15) in node [right] at (0.05,3.15);
    \end{tikzpicture}&
    \begin{tikzpicture}[spy using outlines={rectangle,connect spies}]
      \node[anchor=south west,inner sep=0]  at (0,0) {\includegraphics[width=\mywidthx]{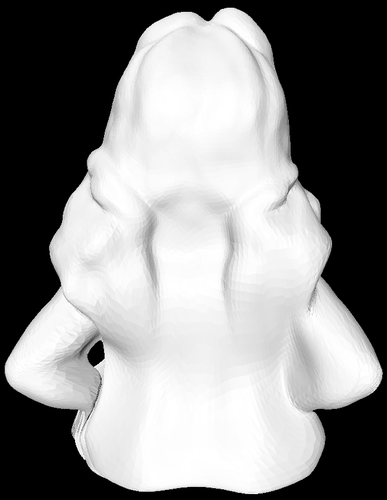} };
      \spy[color=green,width=0.7cm,height=0.9cm,magnification=1.3] on (1.4,2.15) in node [right] at (0.05,3.15);
    \end{tikzpicture}&
    \begin{tikzpicture}[spy using outlines={rectangle,connect spies}]
      \node[anchor=south west,inner sep=0]  at (0,0) {\includegraphics[width=\mywidthx]{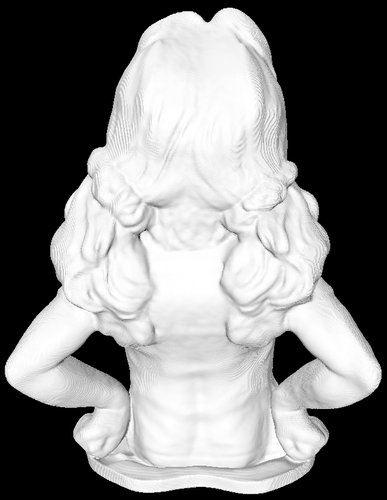} };
      \spy[color=green,width=0.7cm,height=0.9cm,magnification=1.3] on (1.4,2.15) in node [right] at (0.05,3.15);
    \end{tikzpicture}&
    \begin{tikzpicture}[spy using outlines={rectangle,connect spies}]
      \node[anchor=south west,inner sep=0]  at (0,0) {\includegraphics[width=\mywidthx]{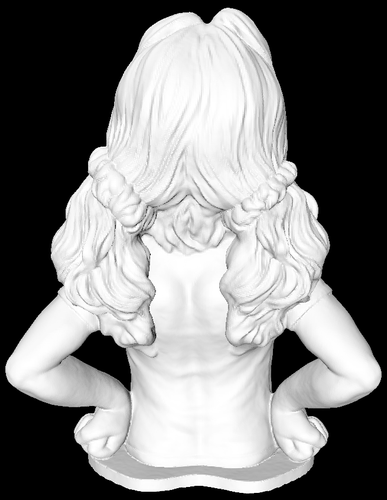}};
      \spy[color=green,width=0.7cm,height=0.9cm,magnification=1.3] on (1.4,2.15) in node [right] at (0.05,3.15);
    \end{tikzpicture}\\
    \begin{tikzpicture}[spy using outlines={rectangle,connect spies}]
      \node[anchor=south west,inner sep=0]  at (0,0) {\includegraphics[width=\mywidthx]{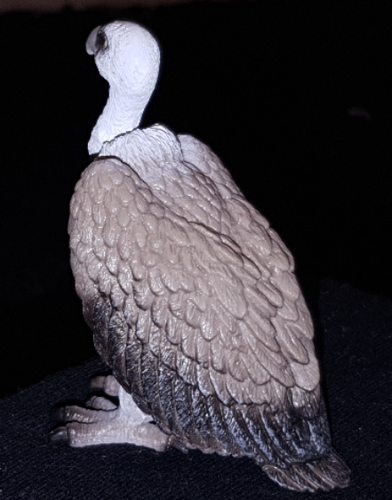}};
      \spy[color=green,width=0.9cm,height=0.95cm,magnification=1.7] on (1.4,1.1) in node [right] at (1.95,3.1);
    \end{tikzpicture}& 
    \begin{tikzpicture}[spy using outlines={rectangle,connect spies}]
      \node[anchor=south west,inner sep=0]  at (0,0) {\includegraphics[width=\mywidthx]{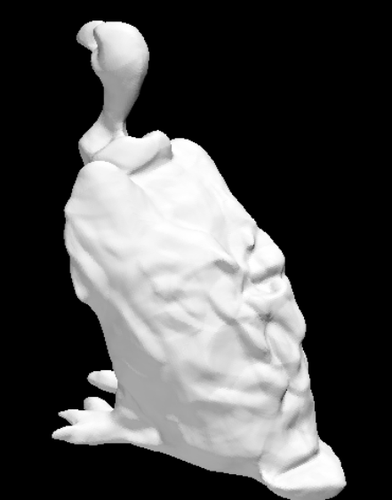}};
      \spy[color=green,width=0.9cm,height=0.95cm,magnification=1.7] on (1.4,1.1) in node [right] at (1.95,3.1);
    \end{tikzpicture}&  
    \begin{tikzpicture}[spy using outlines={rectangle,connect spies}]
      \node[anchor=south west,inner sep=0]  at (0,0) {\includegraphics[width=\mywidthx]{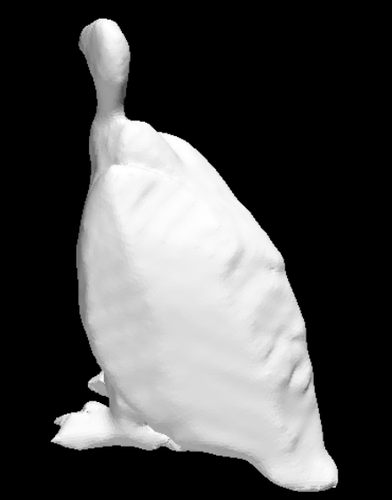}};
      \spy[color=green,width=0.9cm,height=0.95cm,magnification=1.7] on (1.4,1.1) in node [right] at (1.95,3.1);
    \end{tikzpicture}&  
    \begin{tikzpicture}[spy using outlines={rectangle,connect spies}]
      \node[anchor=south west,inner sep=0]  at (0,0) {\includegraphics[width=\mywidthx]{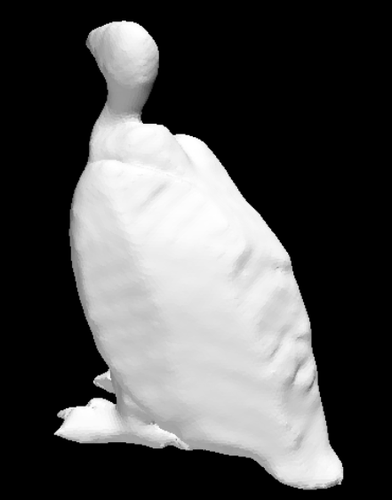}};
      \spy[color=green,width=0.9cm,height=0.95cm,magnification=1.7] on (1.4,1.1) in node [right] at (1.95,3.1);
    \end{tikzpicture}&
    \begin{tikzpicture}[spy using outlines={rectangle,connect spies}]
      \node[anchor=south west,inner sep=0]  at (0,0) {\includegraphics[width=\mywidthx]{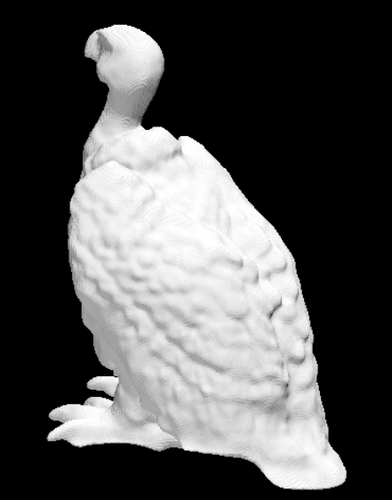}};
      \spy[color=green,width=0.9cm,height=0.95cm,magnification=1.7] on (1.4,1.1) in node [right] at (1.95,3.1);
    \end{tikzpicture}&
    \begin{tikzpicture}[spy using outlines={rectangle,connect spies}]
      \node[anchor=south west,inner sep=0]  at (0,0) {\includegraphics[width=\mywidthx]{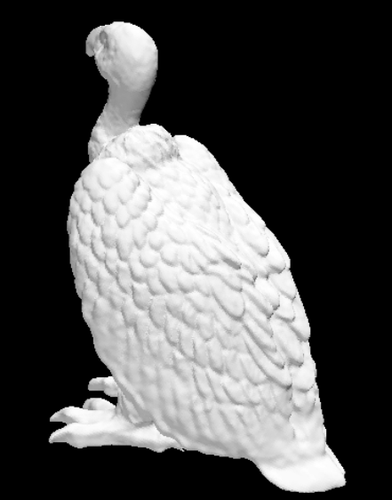}};
      \spy[color=green,width=0.9cm,height=0.95cm,magnification=1.7] on (1.4,1.1) in node [right] at (1.95,3.1);
    \end{tikzpicture}\\
    \begin{tikzpicture}[spy using outlines={rectangle,connect spies}]
      \node[anchor=south west,inner sep=0]  at (0,0) {\includegraphics[width=\mywidthx]{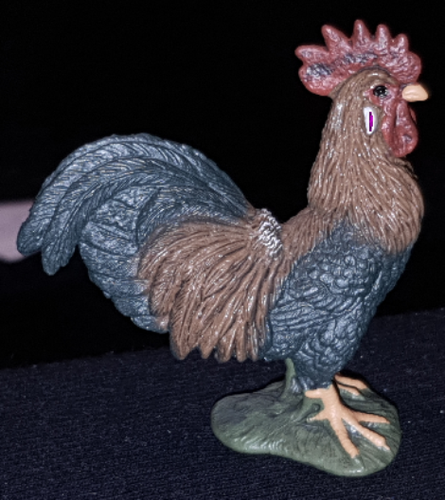}};
      \spy[color=green,width=1.1cm,height=1.1cm,magnification=1.45] on (1.6,1.55) in node [right] at (0.05,0.565);
    \end{tikzpicture}& 
    \begin{tikzpicture}[spy using outlines={rectangle,connect spies}]
      \node[anchor=south west,inner sep=0]  at (0,0) {\includegraphics[width=\mywidthx]{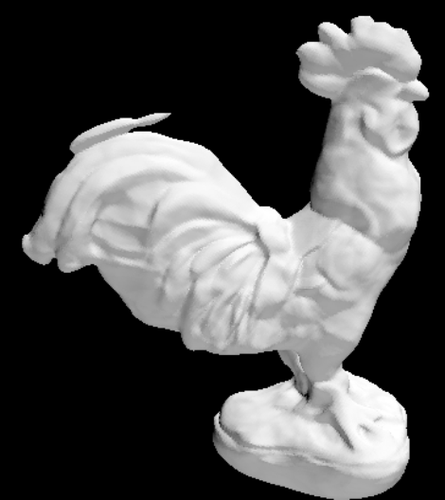}};
      \spy[color=green,width=1.1cm,height=1.1cm,magnification=1.45] on (1.6,1.55) in node [right] at (0.05,0.565);
    \end{tikzpicture}& 
    \begin{tikzpicture}[spy using outlines={rectangle,connect spies}]
      \node[anchor=south west,inner sep=0]  at (0,0) {\includegraphics[width=\mywidthx]{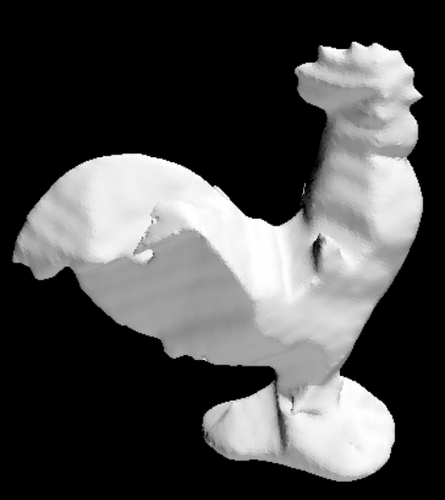}};
      \spy[color=green,width=1.1cm,height=1.1cm,magnification=1.45] on (1.6,1.55) in node [right] at (0.05,0.565);
    \end{tikzpicture}& 
    \begin{tikzpicture}[spy using outlines={rectangle,connect spies}]
      \node[anchor=south west,inner sep=0]  at (0,0) {\includegraphics[width=\mywidthx]{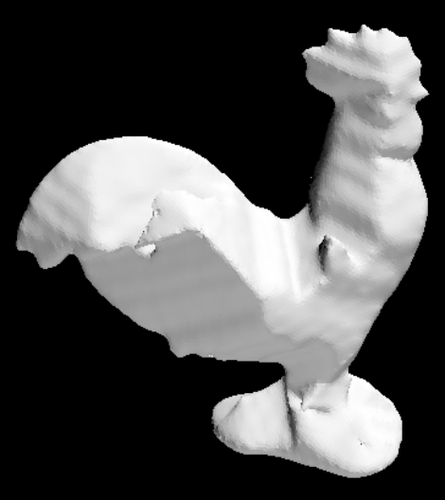}};
      \spy[color=green,width=1.1cm,height=1.1cm,magnification=1.45] on (1.6,1.55) in node [right] at (0.05,0.565);
    \end{tikzpicture}&
    \begin{tikzpicture}[spy using outlines={rectangle,connect spies}]
      \node[anchor=south west,inner sep=0]  at (0,0) {\includegraphics[width=\mywidthx]{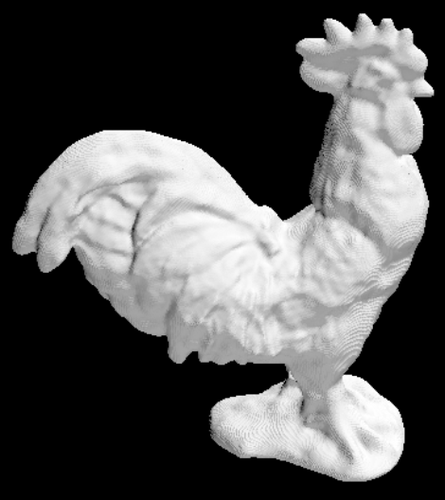} };
      \spy[color=green,width=1.1cm,height=1.1cm,magnification=1.45] on (1.6,1.55) in node [right] at (0.05,0.565);
    \end{tikzpicture}&
    \begin{tikzpicture}[spy using outlines={rectangle,connect spies}]
      \node[anchor=south west,inner sep=0]  at (0,0) {\includegraphics[width=\mywidthx]{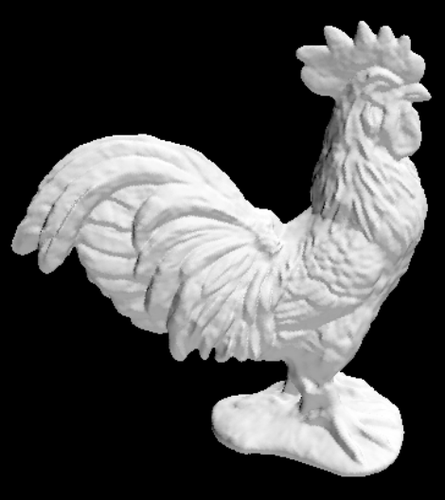}};
      \spy[color=green,width=1.1cm,height=1.1cm,magnification=1.45] on (1.6,1.55) in node [right] at (0.05,0.565);
    \end{tikzpicture}\\
    Input image & \cite{Wu_2023_ICCV} & \cite{cheng2022diffeomorphic} (directional light) & \cite{cheng2022diffeomorphic} (point-light) & \cite{yang2022ps} & Ours
  \end{tabular}
  \xcaption{Full 3D reconstruction of one synthetic and two real objects from 6 viewpoints. While some of the baseline reconstructions (with directional or point-light models) exhibit texture-induced patterns, distortions or other artefacts, the proposed framework consistently yields artefact- and distortion-free reconstructions with finer geometric details. Additional results are shown in the supplementary.}
\label{fig:general_results}
\end{figure*}

\subsection{Using the optimal diffuse albedo}
\label{sec:opt_diffuse_albedo}
In this section, we introduce a %
strategy to handle the diffuse albedo.
This is inspired by~\cite{guo2022edge}, however in their single-view approach they are only able to handle the least-squares case, \ie $L^2$, while we extend their formulation to  objective functions of the form of an $L^p$-norm, with $p=1$ in our case, see \cref{eq:data_term}.
In lieu of employing an MLP for modeling the diffuse albedo, we opt for approximating an optimal solution considering the remaining parameters. This allows us to exclude the diffuse albedo from the optimization process. Our subsequent demonstration will underscore the significant benefits of this approach in the context of sparse views, leading to markedly improved final geometry.
A key technical change is the domain of the diffuse parameter~$\brdfd$. So far, it has been a vector field
defined in world space~$\Omega$, implemented with a neural network.
However, in the following we only try to recover the projection of~$\brdfd$ on the image planes.
Thus, the input to~$\rho$ will be a pixel, and the output is the diffuse albedo at the intersection
between the pixel's ray and the object surface.
In particular, the set of BRDF parameters is reduced to~$\brdfparam=\gamma_2$ and now only includes specular parameters. 

\subheading{Reformulation of the data term}
We will now derive an expression for the data term which allows to solve for
diffuse albedo~$\rho$ when given all of the
other parameters.
Note that the SVBRDF can be decomposed into a diffuse part and a specular part \cite{karis2013real},
\begin{equation}
  \brdf(\px, \pn, \vo, \pl) = \dfrac{\brdfd(\px)}{\pi} + \brdfs(\px,\pn,\vo,\pl),
  \label{eq:brdf_decomposition}
\end{equation}
where the specular part $\brdfs$ depends on both the roughness and specular albedo.
If we denote the BRDF-independent factor of the radiance field~\eqref{eq:point_light_model} by
\begin{equation}
\label{eq:shading}
S^l(\px,\pn) = \dfrac{L_0}{\norm{\px - \plp^l}^2} \max(0,\pn \cdot \pl),
\end{equation}
this implies that the total radiance field is the sum of diffuse radiance~$\frac{\brdfd}{\pi} S^l$
and specular radiance~$L_s^l = \brdfs S^l$.

Inserting this into the volume rendering equation~\eqref{eq:volume_integral},
we can thus re-arrange the data term~\eqref{eq:data_term} as
\begin{equation}
\begin{aligned}
\label{eq:new_data_term}
  E_\text{RGB}(\sdfparam,\phi,\brdfparam,\brdfd) &= \sum_\pp \sum_{l \in S_\tau(\pp)} \norm{b^l_\pp(\sdfparam,\phi,\brdfparam) - \brdfd_\pp a^l_\pp(\sdfparam,\phi)}_1\\
\text{with }
  a^l_\pp(\sdfparam,\phi) &= \frac{1}{\pi}\int_{0}^\infty \pdf_\theta(t) S^l(\px(t),\pn_\theta(t))\dx{t},\\
  b^l_\pp(\sdfparam,\phi,\brdfparam) &= \img^l_\pp - \int_{0}^\infty \pdf_\theta(t) L_s^l(\px(t),\pn_\theta(t),\vo)\dx{t}.
\end{aligned}
\end{equation}
When keeping~$\sdfparam$, $\phi$, and~$\brdfparam$ fixed, this expression can be minimized pixel-wise for $\rho$ as a linear least absolute deviation problem,
as we show next.

\subheading{Alternate treatment of the diffuse albedo}
It is now possible to completely exclude the diffuse albedo $\brdfd$ from the optimization
and simply replace it by the optimal value 
\begin{equation}
\brdfd^*_{\sdfparam,\brdfparam} = \argmin_{\brdfd} E_\text{RGB}(\sdfparam,\phi,\brdfparam,\brdfd)
\label{eq:optimal_albedo}
\end{equation}
with respect to the other parameters.
In our case, we are facing a linear least absolute deviation problem which can be solved individually for each pixel.

\begin{figure*}[t]
  \small
  \hskip 5mm
  \resizebox{0.88\textwidth}{!}{
  \newcommand{\mywidthc}{0.02\textwidth}
  \newcommand{\mywidthx}{0.16\textwidth}
  \newcolumntype{C}{ >{\centering\arraybackslash} m{\mywidthc} }
  \newcolumntype{X}{ >{\centering\arraybackslash} m{\mywidthx} }
  \newcommand{\tabelt}[1]{\hfil\hbox to 0pt{\hss #1 \hss}\hfil}
  \setlength\tabcolsep{1pt} %
  \begin{tabular}{CXXXXX}
    \rotatebox{90}{\dogA{}}&
    \includegraphics[width=\mywidthx]{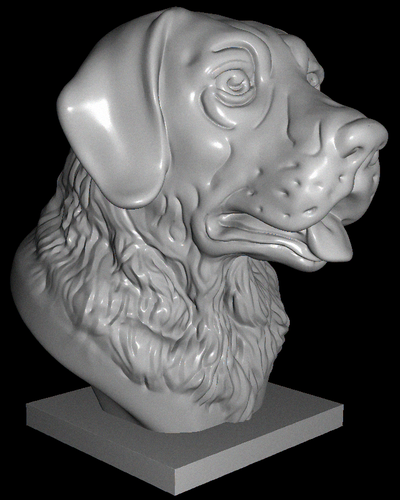}&
    \includegraphics[width=\mywidthx]{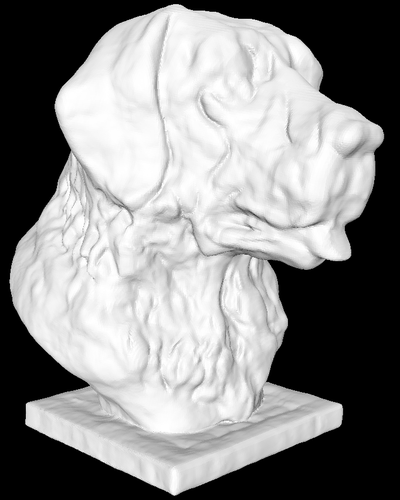} & \includegraphics[width=\mywidthx]{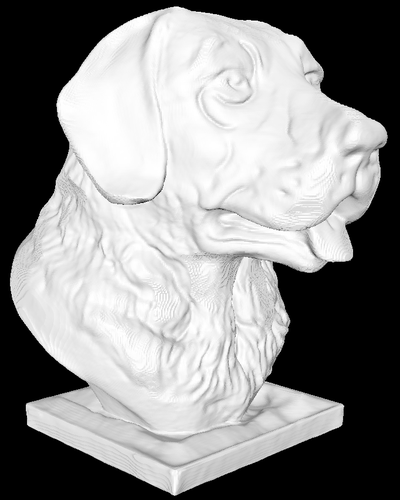} &
    \includegraphics[width=\mywidthx]{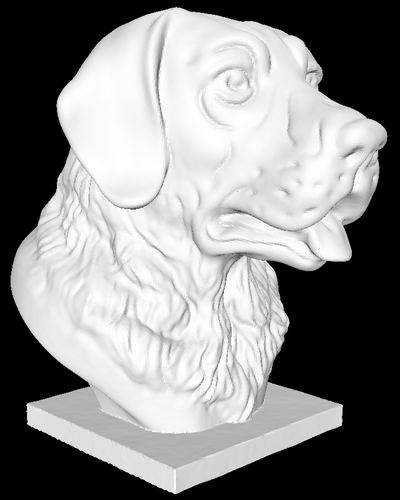}\\
    \rotatebox{90}{\dogB{}}&
    \includegraphics[width=\mywidthx]{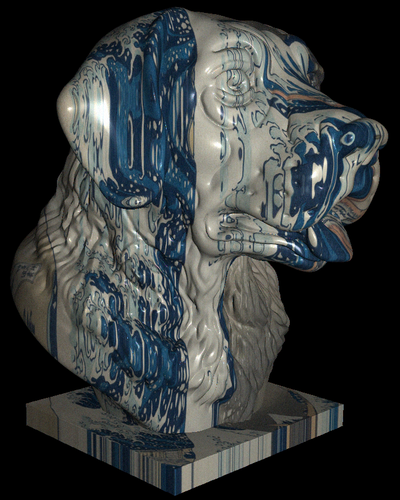}&
    \includegraphics[width=\mywidthx]{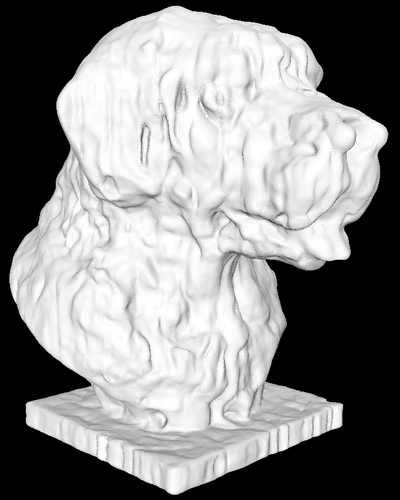} & \includegraphics[width=\mywidthx]{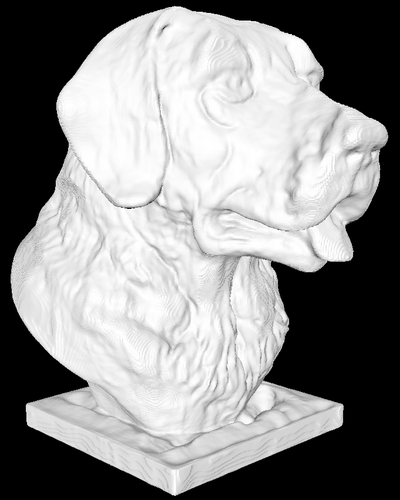} &
    \includegraphics[width=\mywidthx]{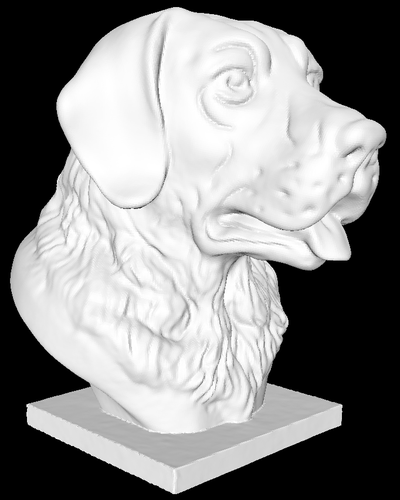}\\
    &Input image & \cite{Wu_2023_ICCV} & \cite{yang2022ps} & Ours
  \end{tabular}
  }
  \vspace{-3mm}
  \caption{
  Results for both \dogA{} and \dogB{} obtained with 5 viewpoints reveal a notable performance degradation in \cite{Wu_2023_ICCV} and \cite{yang2022ps} when confronted with complex textures. This deterioration is attributed to a generalization failure of their learned priors. In contrast, our approach, which does not rely on any prior, provides a superior reconstruction that is unaffected by the complex texture.
  }
  \vspace{-2mm}
\label{fig:texture_dependency}
\end{figure*}

We thus approximate the solution using a few iterations of iterative reweighted least squares~\cite{burrus2012iterative}, as it efficiently solves the problem while providing a differentiable expression. In order to simplify notation, we omit the dependency \wrt the parameters and denote only the channel-wise operations,
where we iteratively compute
\begin{equation}
\brdfd^*_\pp \approx \brdfd^{(k)}_\pp = \dfrac{b_\pp^\top W^{(k)}_\pp a_\pp}{a_\pp^\top W^{(k)}_\pp a_\pp}
\label{eq:optimal_albedo_approx}
\end{equation}
with diagonal matrices
\begin{equation}
W^{(k)}_\pp = 
   \begin{cases}
      \text{diag}( \max( \epsilon, | b_\pp - \brdfd^{(k-1)}_\pp a_\pp | )^{-1}  ) & \text{if } k \geq 1, \\[0.3ex]
        I &\text{otherwise.}
   \end{cases}
\end{equation}
Above, $k$ is the index of iteration, and $a_p$ and $b_p$ are the vectors whose components are given 
in \cref{eq:new_data_term}. The small constant~$\epsilon>0$ avoids numerical issues, and~$I$
is an identity matrix of matching size.
The gradient of $\brdfd^{(k)}_\pp$ with respect to $\sdfparam$ and $\brdfparam$ is computed using automatic differentiation, however the gradient of $W^{(k)}_\pp$ is not back-propagated, as it makes the optimization much more difficult and degrades the quality of the results.

In the forthcoming evaluation section, we demonstrate that this formulation delivers exceptional performance in $3$D reconstruction from sparse viewpoints. Notably, it proves particularly advantageous in the most sparse scenarios, where only two viewpoints are available, showcasing its superiority over the previous MLP-based formulation.

\section{Results}
\vspace{-2mm}
\label{sec:Results}
\begin{table}[t]
\footnotesize
\setlength{\tabcolsep}{2pt}
\resizebox{1.0\linewidth}{!}{
\begin{tabular}{l|ccccc|ccccc|}
~ & \multicolumn{5}{c|}{$\downarrow$RMSE$\times 100$} & \multicolumn{5}{c|}{$\downarrow$MAE} \\
~ & \cite{Wu_2023_ICCV} & \cite{cheng2022diffeomorphic}D & \cite{cheng2022diffeomorphic}P & \cite{yang2022ps} & Ours & \cite{Wu_2023_ICCV} & \cite{cheng2022diffeomorphic}D & \cite{cheng2022diffeomorphic}P & \cite{yang2022ps} & Ours\\
\cline{1-11}
& & & & & & & & & &\\[-2mm]
\dogA{} & 2.1 & 8.9 & 13.9 & 2.1 & \textbf{0.4} & 22.0 & 39.9 & 39.2 & 18.8 & {\color{white}0}\textbf{4.6} \\
\dogB{} & 2.2 & 8.4 & 13.4 & 2.1 & \textbf{0.4} & 23.2 & 29.6 & 40.3 & 19.5 & {\color{white}0}\textbf{4.8}\\
\girlA{} & 0.9 & 6.1 & {\color{white}0}2.3 & 1.5 & \textbf{0.3} & 17.2 & 41.8 & 46.5 & 23.8 & {\color{white}0}\textbf{9.6}\\
\girlB{} & 1.5 & 5.2 & {\color{white}0}4.8 & 1.5 & \textbf{0.3} & 23.9 & 40.0 & 47.3 & 23.4 & \textbf{10.2}
\end{tabular}
}
\xcaption{RMSE and MAE for full 3D reconstructions. RMSE is computed based on the vertex to mesh distance, and the MAE is computed using the angular error between the normals of a vertex and its closest point in the ground truth mesh. \OurAlbedoNet{} leads to highly similar results as our main framework in this scenario.}
\vspace{-2mm}
\label{tab:main_results}
\end{table}
\begin{figure*}[t]
  \small
  \hskip 2mm
  \resizebox{0.92\textwidth}{!}{
  \newcommand{\mywidthc}{0.02\textwidth}
  \newcommand{\mywidthx}{0.15\textwidth}
  \newcolumntype{C}{ >{\centering\arraybackslash} m{\mywidthc} }
  \newcolumntype{X}{ >{\centering\arraybackslash} m{\mywidthx} }
  \newcommand{\tabelt}[1]{\hfil\hbox to 0pt{\hss #1 \hss}\hfil}
  \setlength\tabcolsep{1pt} %
  \begin{tabular}{CXXXXX}
    \rotatebox{90}{\squirrel{}}&
    \includegraphics[width=\mywidthx]{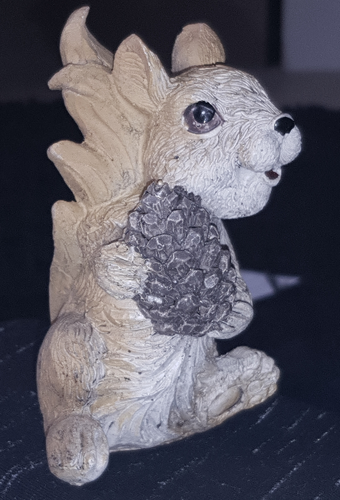}&
    \includegraphics[width=\mywidthx]{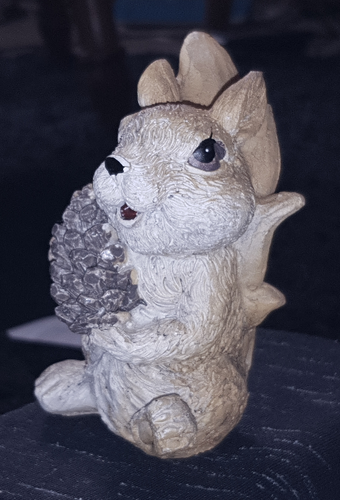} & \includegraphics[width=\mywidthx]{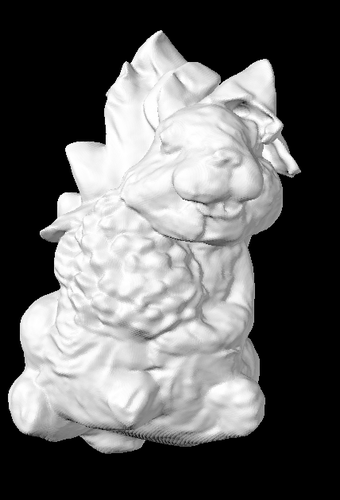} &
    \includegraphics[width=\mywidthx]{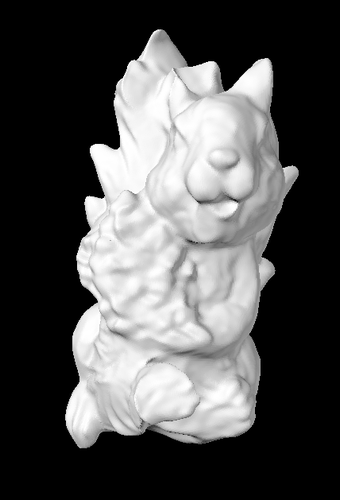} &
    \includegraphics[width=\mywidthx]{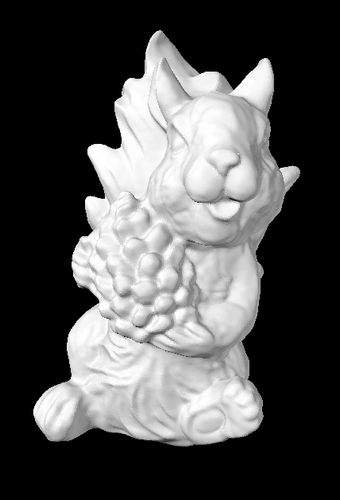}\\
    \rotatebox{90}{\bird{}}&
    \includegraphics[width=\mywidthx]{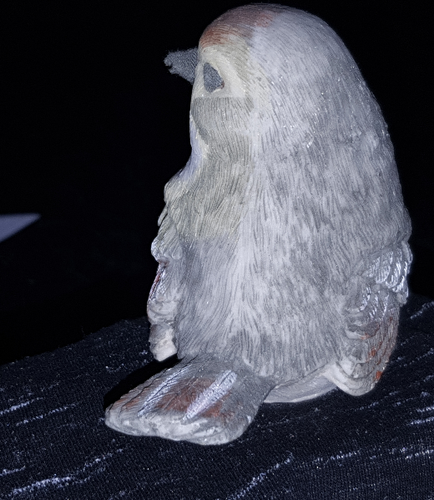}&
    \includegraphics[width=\mywidthx]{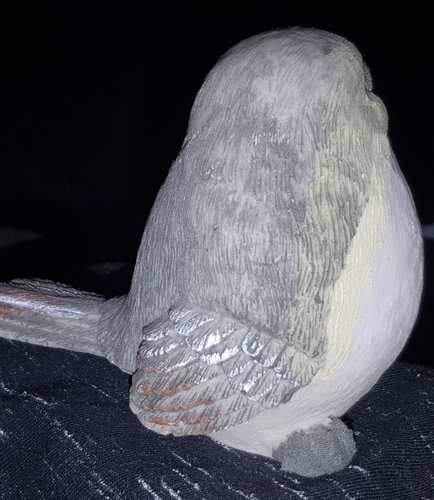} & \includegraphics[width=\mywidthx]{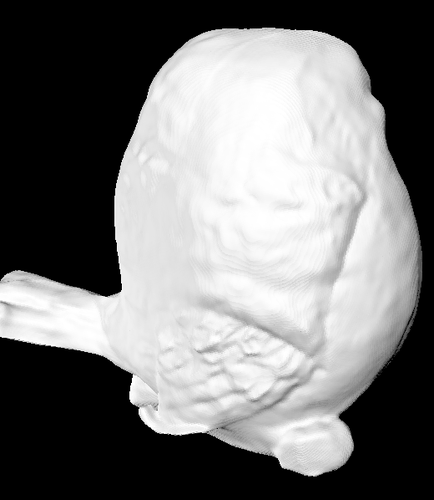} &
    \includegraphics[width=\mywidthx]{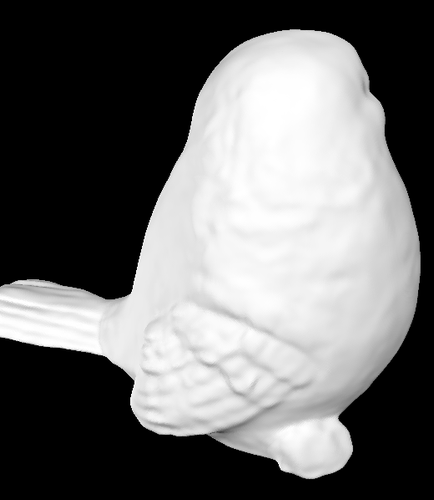} &
    \includegraphics[width=\mywidthx]{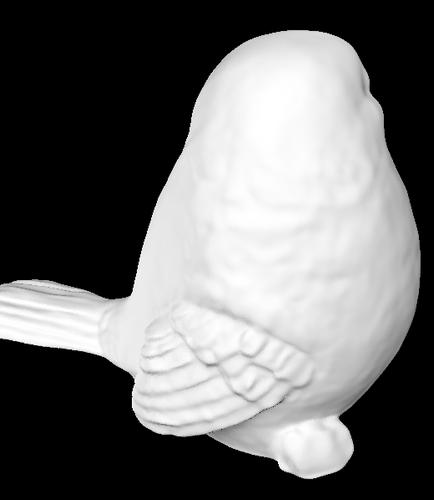}\\
    \rotatebox{90}{\dogB{}}&
    \includegraphics[width=\mywidthx]{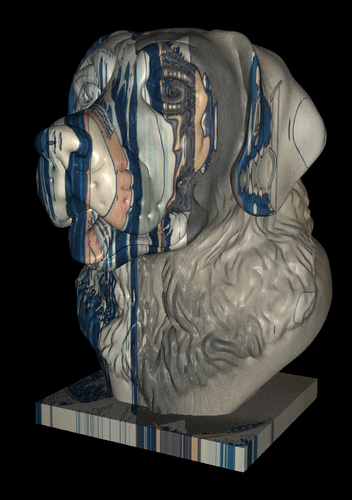}&
    \includegraphics[width=\mywidthx]{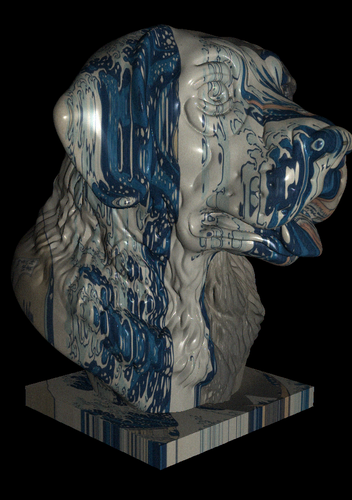} & \includegraphics[width=\mywidthx]{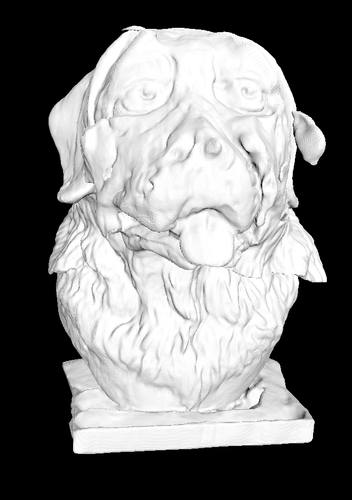} &
    \includegraphics[width=\mywidthx]{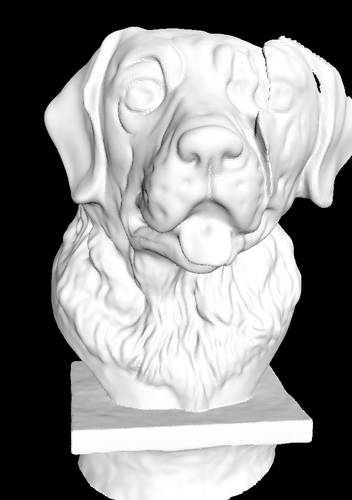} &
    \includegraphics[width=\mywidthx]{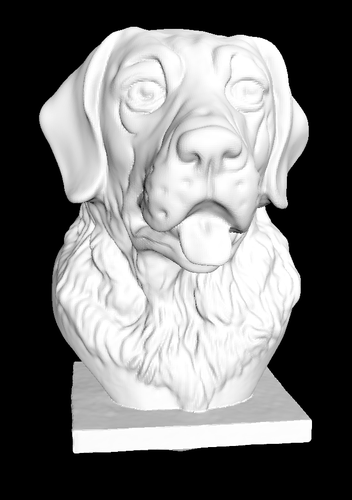}\\
    &Viewpoint 1 & Viewpoint 2 & \cite{yang2022ps} & OurAlbedoNet & Ours
  \end{tabular}
  }%
  \vspace{-1mm}
  \xcaption{Reconstructions from two distant viewpoints. Severe artefacts can be noticed for both \cite{yang2022ps} and \OurAlbedoNet, and more particularly \cite{yang2022ps} introduce significant distortions in the shapes, as a result of using normal maps which do not provide strong enough constraints on the absolute position of the points. }
\vspace{-2mm}
\label{fig:two_viewpoints}
\end{figure*}

To substantiate the efficacy of our framework, we conduct evaluations on both synthetic and real-world datasets. We generate four synthetic scans by combining two distinct geometries with two different materials. The first material is white, rendering it textureless, while the second material exhibits a high degree of texture. This design enables us to quantify the influence of texture on the obtained results. The four synthetic scans are denoted as \dogA{}, \dogB{}, \girlA{}, and \girlB{}. Additionally, we acquire real-world data for six objects, namely, \bird{}, \squirrel{}, \hawk{}, \rooster{}, \flamingo{} and \pumpkin{}. We also consider the DiLiGenT-MV dataset \cite{li2020multi}.\\[2.0mm]
\subheading{Evaluation} We first evaluate full 3D reconstruction using only six viewpoints in total, except for \dogA{} and \dogB{} where five viewpoints are considered. We evaluate against both state-of-the-art sparse viewpoint reconstruction approaches assuming static illumination \cite{long2022sparseneus,ren2023volrecon,liang2023retr,Wu_2023_ICCV} and photometric stereo approaches \cite{cheng2022diffeomorphic,yang2022ps}. Since \cite{cheng2022diffeomorphic} allows for directional and point-light illumination, %
we consider both cases in our evaluation. 
\cite{cheng2022diffeomorphic}D and \cite{cheng2022diffeomorphic}P refers to the directional and point-light version, respectively. 
Given the calibrated nature of their framework, we furnish the estimated lighting parameters using our approach. Additionally, we explore a more challenging scenario wherein only two viewpoints with a wide baseline are available. We therefore introduce an ablated version of our framework, wherein the diffuse albedo is modeled with an MLP. This adaptation allows to highlight the advantages of employing the optimal diffuse albedo introduced in \cref{sec:opt_diffuse_albedo}.\\[2.0mm]
\subheading{Full 3D reconstruction}
As anticipated, \cite{long2022sparseneus, ren2023volrecon, liang2023retr} produce degenerate meshes in our wide baseline setup. Results in the supplementary demonstrate that they perform adequately when the distance between cameras is sufficiently small, although our framework consistently outperforms them. Conversely, \cref{tab:main_results} and \cref{fig:general_results} illustrate that our approach surpasses relevant baselines \cite{Wu_2023_ICCV, cheng2022diffeomorphic, yang2022ps} both quantitatively and qualitatively. Our method yields sharper details without visible artifacts. Furthermore, error maps (shown in the supplementary material) reveal that the geometry obtained with \cite{yang2022ps} exhibits global distortions. Additionally, \cref{fig:texture_dependency} demonstrates that both \cite{Wu_2023_ICCV, yang2022ps} perform poorly when confronted with complex textures, while our results remain unaffected. This suggests that their pre-trained priors failed to generalize for the given texture, possibly misassigning it as geometric patterns. This shows the advantage of approaches without relying on learned priors.
The results for the DiLiGenT-MV dataset \cite{li2020multi} are shown in the supplementary, demonstrating a high accuracy in the classical scenario of a dark room and distant light sources.\\[2.0mm] %
\subheading{Two viewpoints} We denote our ablated framework, which uses a diffuse albedo network, as \OurAlbedoNet{}. We exclude \cite{Wu_2023_ICCV} from consideration as they require at least three viewpoints. \cref{fig:two_viewpoints} vividly illustrates the advantage of our framework in this challenging scenario, especially with the use of the optimal diffuse albedo. Notably, \cite{yang2022ps} and \OurAlbedoNet{} exhibit pronounced artifacts and inaccuracies in geometric details in certain regions. Moreover, \cite{yang2022ps} results in severe distortions of the shape. We attribute this to their heavy reliance on per-view normal maps for reconstruction, which lacks absolute depth information about the shape.
In contrast, our method directly estimates the shape from various point-light illuminations, coupled with a suitable model (\cref{eq:point_light_model}). This approach provides relevant cues about the absolute position of the points, contributing to more accurate and high-quality reconstructions.\\[1mm]
\subheading{Limitations and future work}
A dedicated section is available in the supplementary.\vspace{-2mm}

\section{Conclusion}
\vspace{-1.5mm}
\label{sec:Conclusion}
We introduced the first framework for multi-view uncalibrated point-light photometric stereo. It combines a state-of-the-art volume rendering approach with a physically realistic illumination model consisting of an ambient component and uncalibrated point-lights. As demonstrated in a variety of experiments, the proposed approach offers a practical paradigm to create highly accurate $3$D reconstructions from sparse and distant viewpoints, even outside a controlled dark room environment.\\[0.5mm]
\textbf{Acknowledgment}
This work was supported by the ERC Advanced Grant SIMULACRON, and the Deutsche Forschungsgemeinschaft (DFG, German Research Foundation) under Germany's Excellence Strategy – EXC 2117 – 422037984.

\newpage
{
    \small
    \bibliographystyle{ieeenat_fullname}
    \bibliography{biblio}
}

\newpage

\setcounter{figure}{0}
\appendix
\maketitlesupplementary
\begin{abstract}
   In this supplementary material, we show further details about our framework. Specifically, we describe the network architecture with all its parameters and training specifications. Then we elaborate on the capturing process to retrieve the synthetic and real-world photometric images. After that, we show additional results on three viewpoints with a small camera baseline, as well as error maps of the reconstructions presented in the main paper. Furthermore, we show additional reconstruction results on both captured scans and the multiview diligent dataset, as well as some relighting results. We also analyze the effect of the ratio of point light intensity on the reconstruction quality, as well as the effect of the number of viewpoints and lights. Finally, we elaborate on the limitations of our approach.
\end{abstract}

\section{Network Details}

\subsection{Architecture}
As mentioned in the main paper, we use two multilayer perceptrons (MLPs). The first one describes the geometry via an SDF, $\sdf_\sdfparam$, and the other one is used for the specular parameters of the material, $\alpha_{\brdfparam}$.
The MLP of $\sdf_\sdfparam$ consists of $6$ layers of width $256$, with a skip connection at the $4$-th layer, while the MLP $\alpha_{\brdfparam}$ consist of $3$ layers of width $256$.\\
In order to compensate the spectral bias of MLPs \cite{mildenhall2021nerf}, the input is encoded by positional encoding using $6$ frequencies for both $\sdf_\sdfparam$ and $\alpha_{\brdfparam}$.
For the ablation \OurAlbedoNet, a third MLP describing the BRDF's diffuse albedo, $\brdfd_{\diffuseparam}$, is considered. It consists of $4$ layers of width $512$, and the input is encoded by positional encoding using $12$ frequencies.
\subsection{Parameters and Cost Function}
Similarly to \cite{brahimi2022supervol,iron-2022,yariv2020multiview}, we assume that the scene of interest lies within the unit sphere, which can be achieved by normalizing the camera positions appropriately.
To approximate the Volume rendering integral (4) using (5), we use $m=98$ samples which are also used to approximate (3), all with the sampling strategy of~\cite{yariv2021volume}.\\
We set the objective's function trade-off parameters $\lambda_1 = \lambda_2=0.1$. Furthermore, the terms of the objective function (7) and (8) consist of a batch size of $800$ (inside the silhouette) and $1000$, respectively.
For the mask term (9), we use the same batch as (7) and add $900$ additional rays outside the silhouette whose rays still intersect with the unit sphere.\\
Finally, we always normalize each objective function's summand with its corresponding batch size.
\subsection{Training}
Our networks are trained using the Adam optimizer~\cite{kingma2014adam} with a learning rate initialized with $5\e-4$ and decayed exponentially during training to $5\e-5$, except for the MLP $\alpha_{\brdfparam}$ whose learning rate is constantly equal to $1\e-5$. The light positions $\phi$ are initialized with the camera position of their corresponding viewpoint, with a learning rate initialized with $1\e-2$, and decayed exponentially with the same rate as the other networks.
The remaining parameters are kept to Pytorch's default.\\
We train for $800$ epochs, which lasts about 6 hours using a single NVIDIA Titan GTX GPU with $12$GB memory and $6$ viewpoints.

\section{Data Acquisition}
In this section, we describe how we generated the datasets used in this paper.
\subsection{Synthetic Data}
The synthetic datasets \dogA, \dogB, \girlA, \girlB{} were generated using Blender \cite{blender} and Matlab \cite{MATLAB:2020}, where Blender \cite{blender} is used to render normal, depth and BRDF parameter maps for each viewpoint, and Matlab \cite{MATLAB:2020} is used to render images using equation (1) of the main paper. We used $20$ point light illuminations for each viewpoint, with a ratio of $70\%$ of point light intensity (thus $30\%$ of ambient light), and we also added a zero-mean Gaussian noise with a standard deviation $\sigma=0.02$.
\subsection{Real World Data}
In order to generate the real-world datasets \squirrel{}, \bird{}, \hawk{}, \rooster{}, \flamingo{} and \pumpkin{}, we used a Samsung Galaxy Note 8 and the application "CameraProfessional"\footnote{\url{https://play.google.com/store/apps/details?id=com.azheng.camera.professional}} to generate RAW images as well as the smartphone's images in parallel.
We use the RAW images for our algorithm, and we pre-processed those using Matlab \cite{MATLAB:2020} by following \cite{sumner2014processing}. Since our approach assumes very precise camera parameters, and in order to facilitate calibration, we captured a higher amount of viewpoints and used COLMAP~\cite{schonberger2016pixelwise} to obtain both camera poses and intrinsics with the smartphone's images.\\
We move a hand-held LED\footnote{We use white LUXEON Rebel LED: \url{https://luxeonstar.com/product-category/led-modules/}} to obtain $20$ images with different point light illumination per viewpoint.
\vspace{-0.5mm}
\section{Small camera baseline}

\begin{figure*}[t]
  \small
  \newcommand{\mywidthc}{0.02\textwidth}
  \newcommand{\mywidthn}{0.08\textwidth}
  \newcommand{\mywidthx}{0.164\textwidth}
  \newcolumntype{C}{ >{\centering\arraybackslash} m{\mywidthc} }
  \newcolumntype{N}{ >{\centering\arraybackslash} m{\mywidthn} }
  \newcolumntype{X}{ >{\centering\arraybackslash} m{\mywidthx} }
  \newcommand{\tabelt}[1]{\hfil\hbox to 0pt{\hss #1 \hss}\hfil}
  \setlength\tabcolsep{1pt} %
  \begin{tabular}{CcXXXX}
    \rotatebox{90}{\girlA{}} &
    \begin{tabular}{cc}
    \begin{tabular}{c}
    \includegraphics[width=\mywidthn]{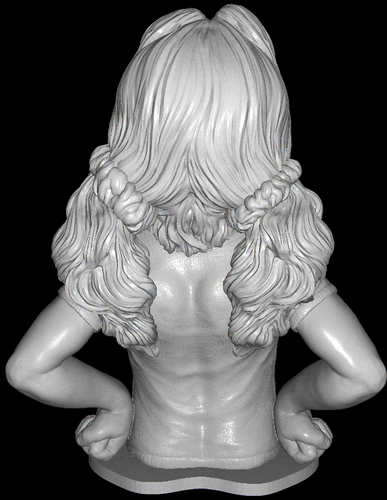}
    \end{tabular}
     &
    \begin{tabular}{c}
    \includegraphics[width=\mywidthn]{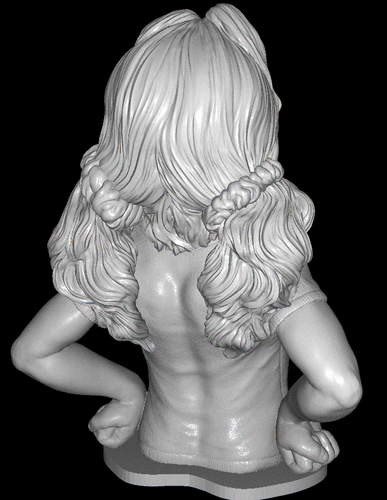}\\
    \includegraphics[width=\mywidthn]{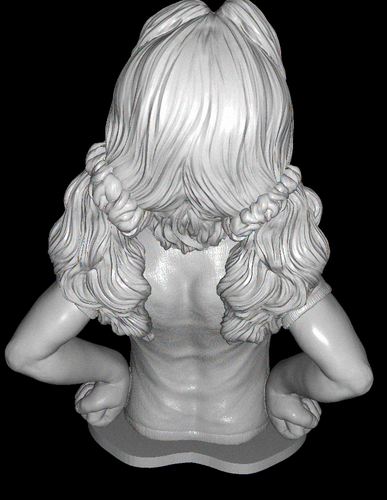}
    \end{tabular} 
    \end{tabular} &
    \includegraphics[width=\mywidthx]{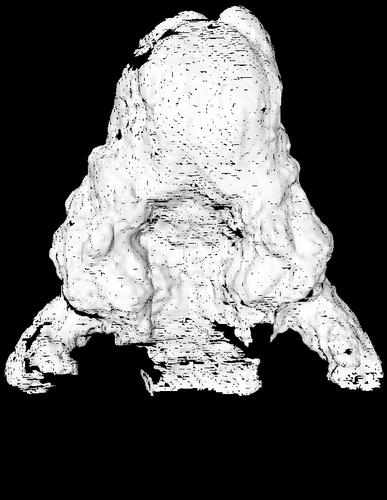}&  
    \includegraphics[width=\mywidthx]{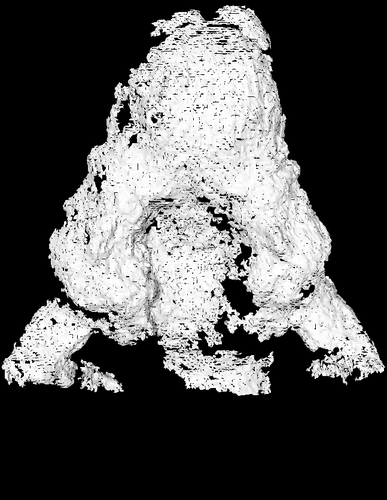}&
    \includegraphics[width=\mywidthx]{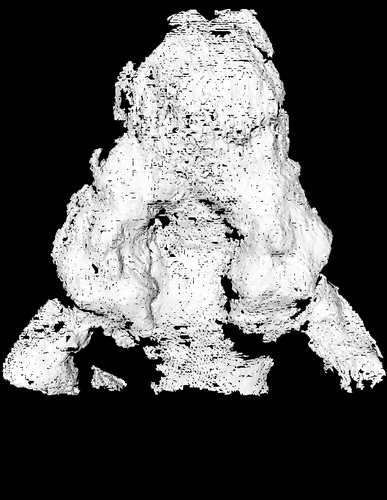}&
    \includegraphics[width=\mywidthx]{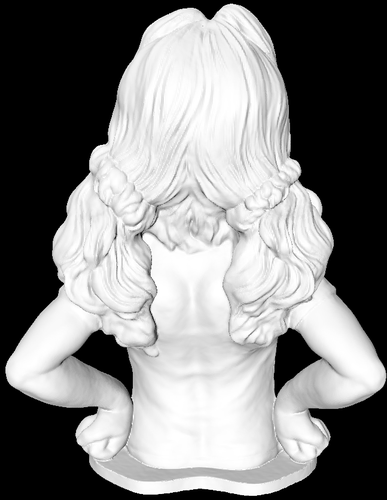}\\
     \rotatebox{90}{\girlB{}} &
    \begin{tabular}{cc}
    \begin{tabular}{c}
    \includegraphics[width=\mywidthn]{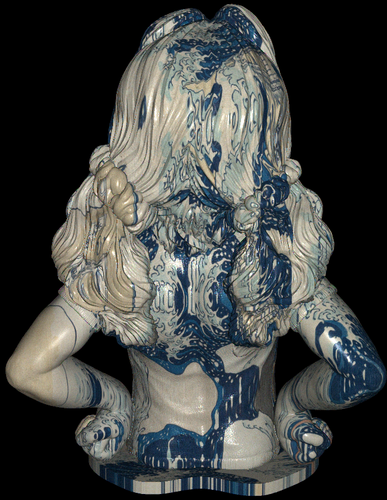}
    \end{tabular}
     &
    \begin{tabular}{c}
    \includegraphics[width=\mywidthn]{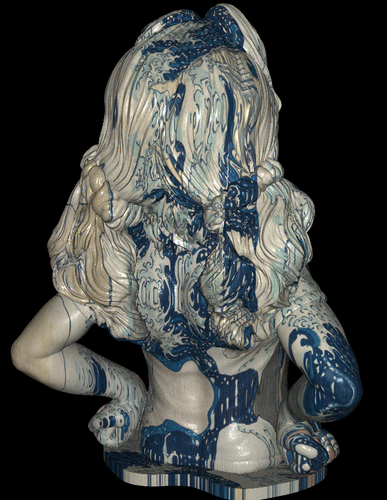}\\
    \includegraphics[width=\mywidthn]{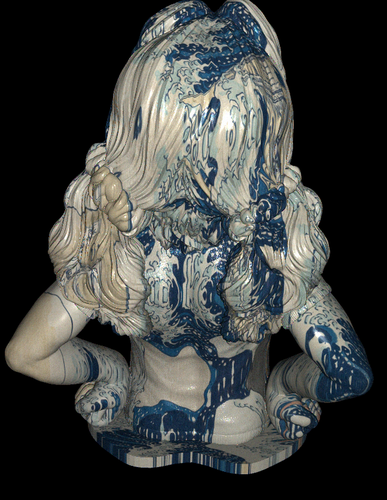}
    \end{tabular} 
    \end{tabular} &
    \includegraphics[width=\mywidthx]{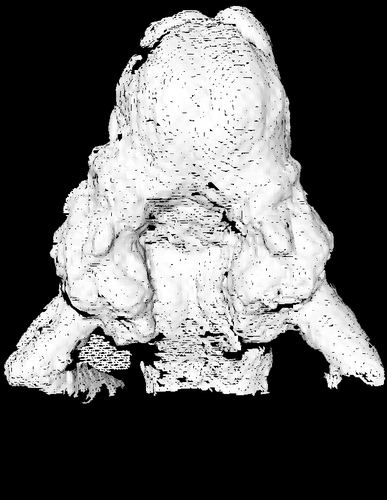}&  
    \includegraphics[width=\mywidthx]{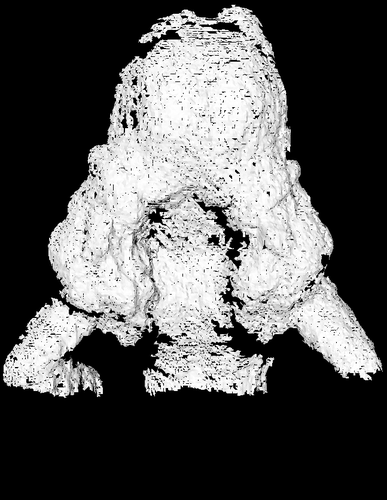}&
    \includegraphics[width=\mywidthx]{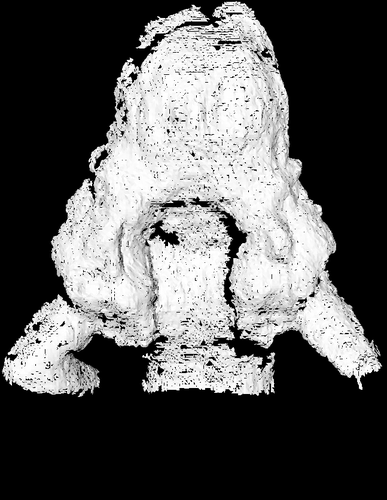}&
    \includegraphics[width=\mywidthx]{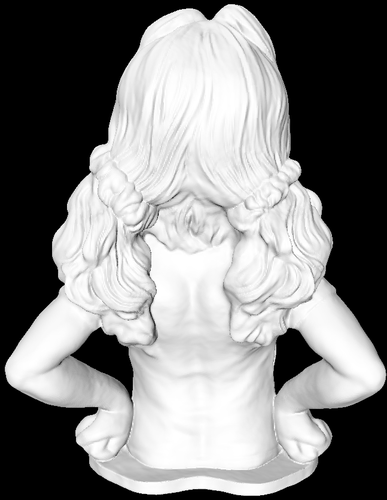}\\
     \rotatebox{90}{\dogA{}} &
    \begin{tabular}{cc}
    \begin{tabular}{c}
    \includegraphics[width=\mywidthn]{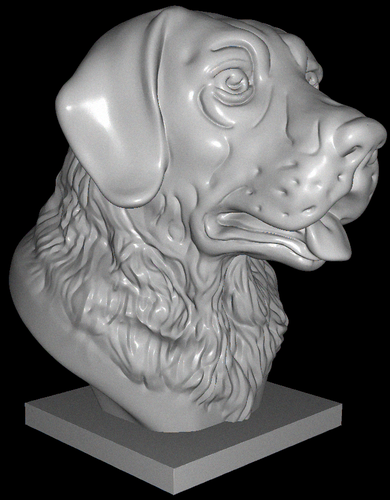}
    \end{tabular}
     &
    \begin{tabular}{c}
    \includegraphics[width=\mywidthn]{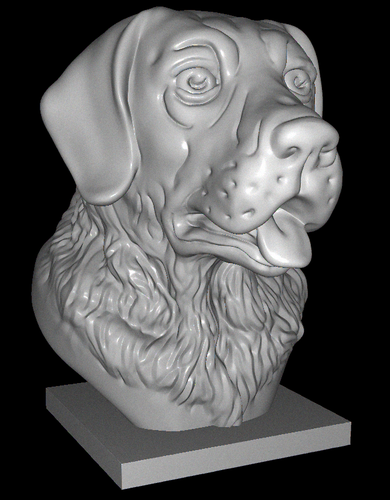}\\
    \includegraphics[width=\mywidthn]{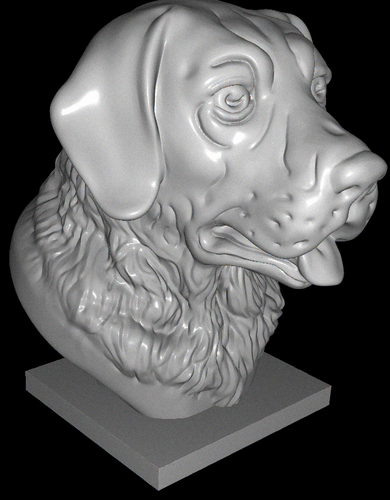}
    \end{tabular} 
    \end{tabular} &
    \includegraphics[width=\mywidthx]{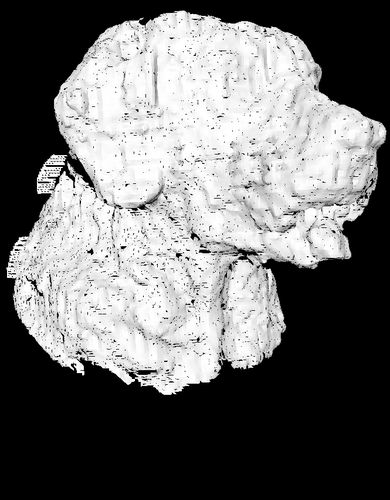}&  
    \includegraphics[width=\mywidthx]{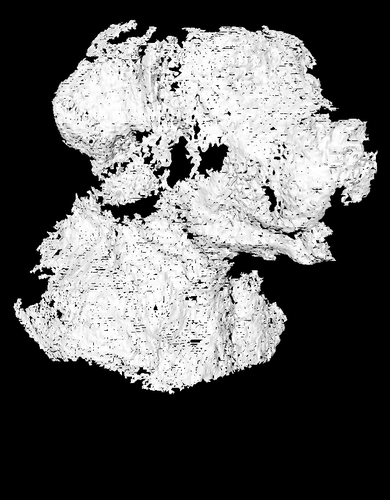}&
    \includegraphics[width=\mywidthx]{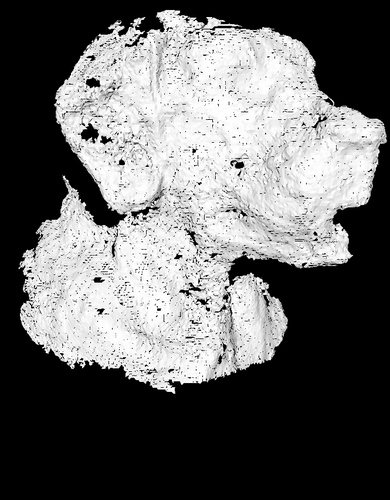}&
    \includegraphics[width=\mywidthx]{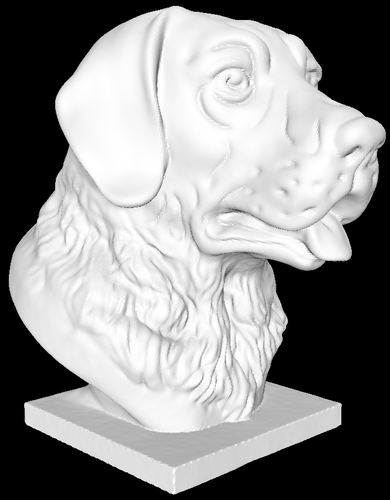}\\
     \rotatebox{90}{\dogB{}} &
    \begin{tabular}{cc}
    \begin{tabular}{c}
    \includegraphics[width=\mywidthn]{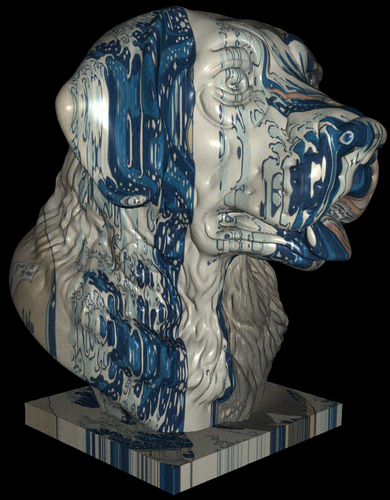}
    \end{tabular}
     &
    \begin{tabular}{c}
    \includegraphics[width=\mywidthn]{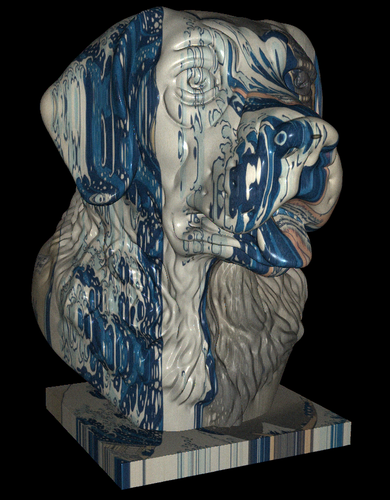}\\
    \includegraphics[width=\mywidthn]{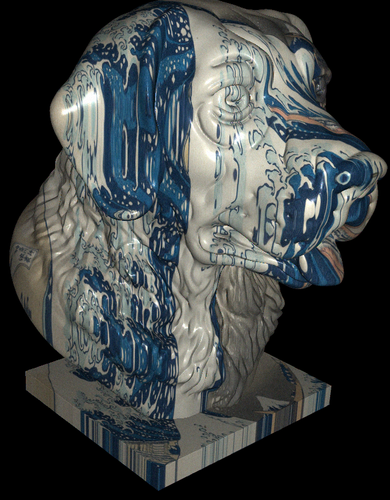}
    \end{tabular} 
    \end{tabular} &
    \includegraphics[width=\mywidthx]{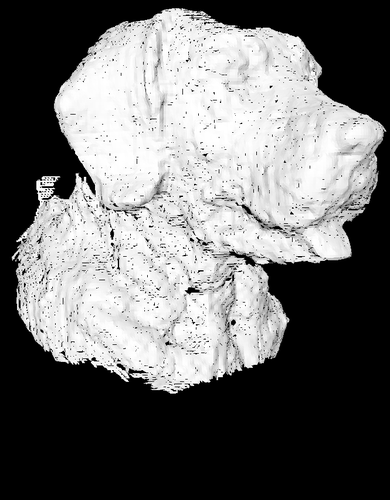}&  
    \includegraphics[width=\mywidthx]{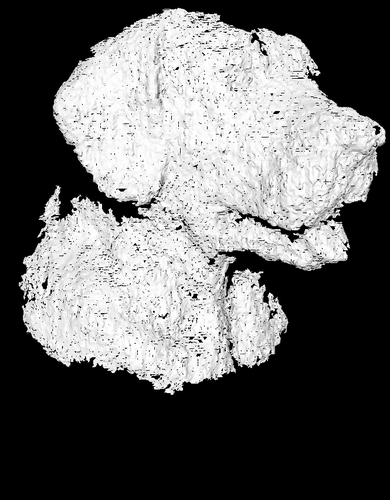}&
    \includegraphics[width=\mywidthx]{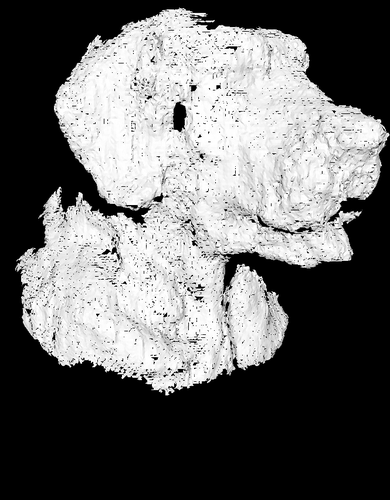}&
    \includegraphics[width=\mywidthx]{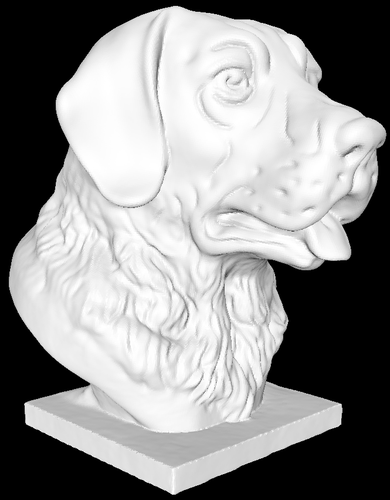}\\
    & Input viewpoints & \cite{long2022sparseneus} & \cite{ren2023volrecon} & \cite{liang2023retr} & Ours
  \end{tabular}
  \xcaption{Results using three viewpoints with small camera baseline.}
\label{fig:three_viewpoints}
\end{figure*}

As mentioned in the main paper, \cite{long2022sparseneus,ren2023volrecon,liang2023retr} lead to degenerate meshes when considering distant cameras, and are thus not suited to reconstruct full $3$D objects from sparse viewpoints. For a more fair comparison, we focus here on a different scenario, where only a part of the object is reconstructed from three viewpoints with a very small camera baseline. \cref{fig:three_viewpoints} clearly indicates that our approach allows for much more accurate and complete $3$D reconstruction than \cite{long2022sparseneus,ren2023volrecon,liang2023retr}. Note that all the meshes were obtained solely by using the official implementations.

\vspace{-0.5mm}
\section{Error maps}
\begin{figure*}[t]
  \small
  \newcommand{\mywidthc}{0.02\textwidth}
  \newcommand{\mywidthx}{0.164\textwidth}
  \newcolumntype{C}{ >{\centering\arraybackslash} m{\mywidthc} }
  \newcolumntype{X}{ >{\centering\arraybackslash} m{\mywidthx} }
  \newcommand{\tabelt}[1]{\hfil\hbox to 0pt{\hss #1 \hss}\hfil}
  \setlength\tabcolsep{1pt} %
\begin{tabular}{cc}
  \begin{tabular}{cXXXXX}
    \rotatebox{90}{\girlA{}}&
    \includegraphics[width=\mywidthx]{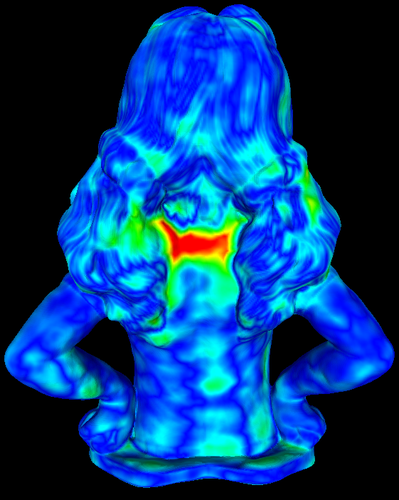} &
    \includegraphics[width=\mywidthx]{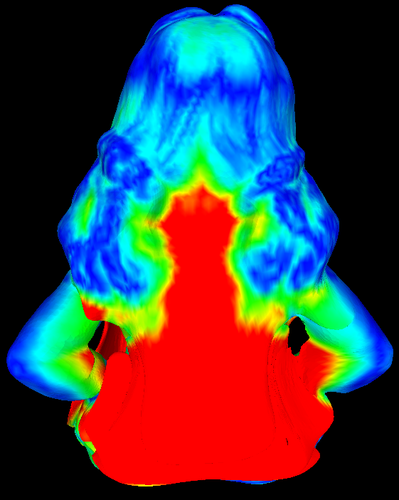}&  
    \includegraphics[width=\mywidthx]{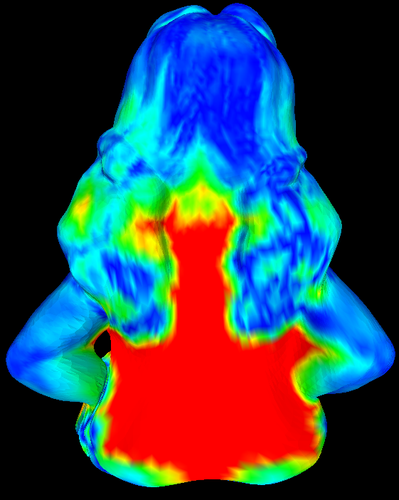}&
    \includegraphics[width=\mywidthx]{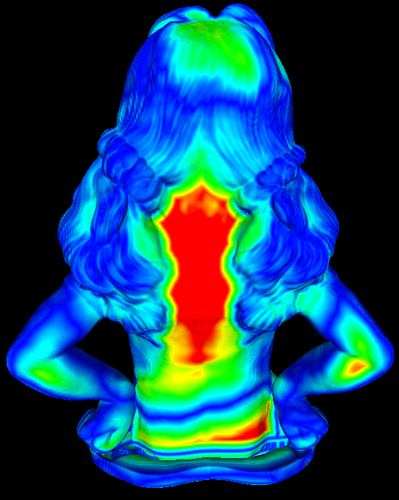}&
    \includegraphics[width=\mywidthx]{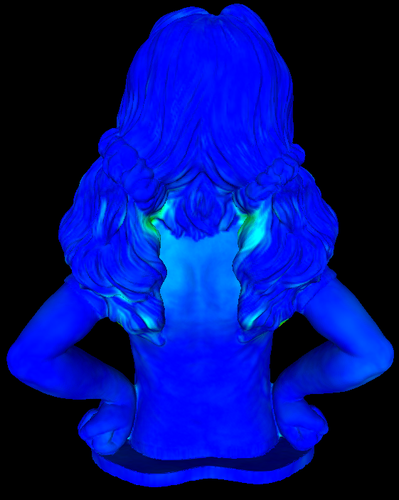}\\
    \rotatebox{90}{\girlB{}}&
    \includegraphics[width=\mywidthx]{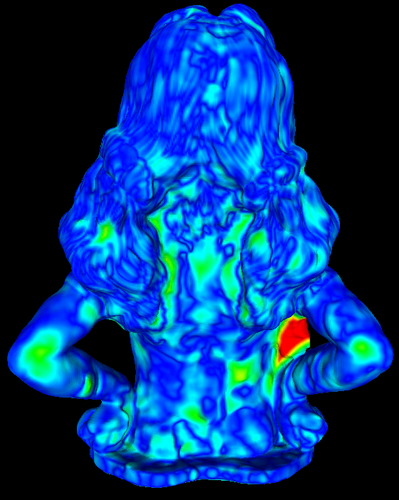} &
    \includegraphics[width=\mywidthx]{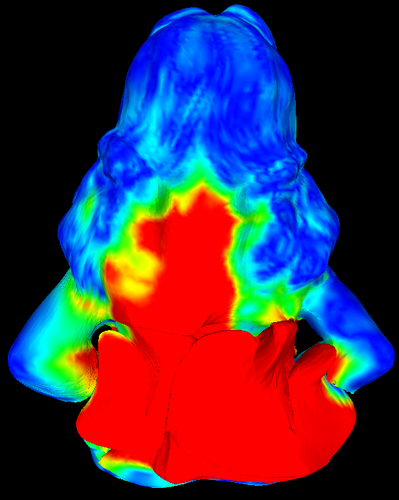}&  
    \includegraphics[width=\mywidthx]{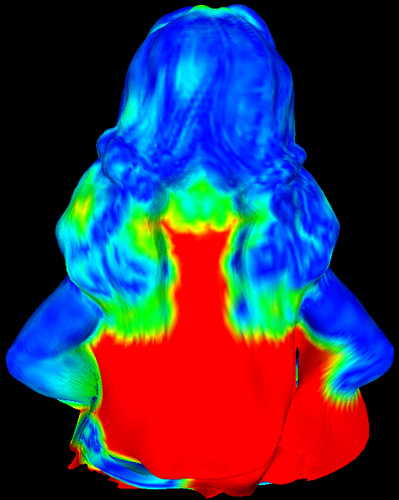}&
    \includegraphics[width=\mywidthx]{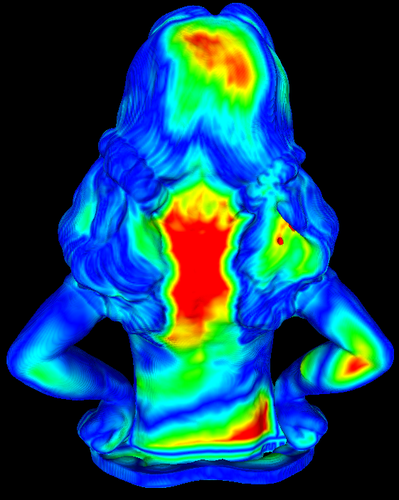}&
    \includegraphics[width=\mywidthx]{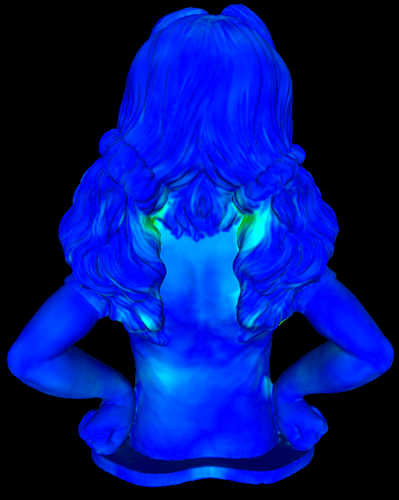}\\
    \rotatebox{90}{\dogA{}}&
    \includegraphics[width=\mywidthx]{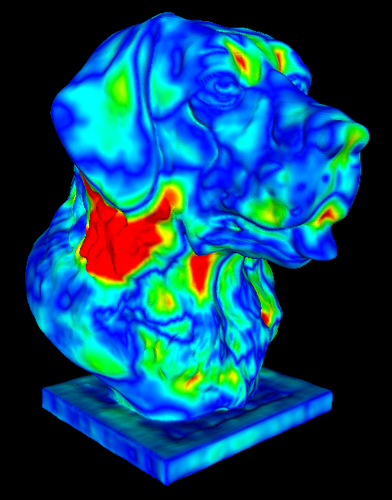} &
    \includegraphics[width=\mywidthx]{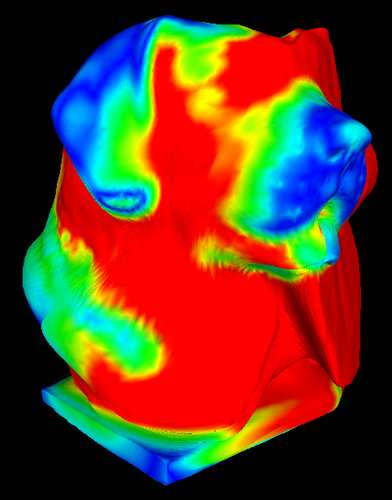}&  
    \includegraphics[width=\mywidthx]{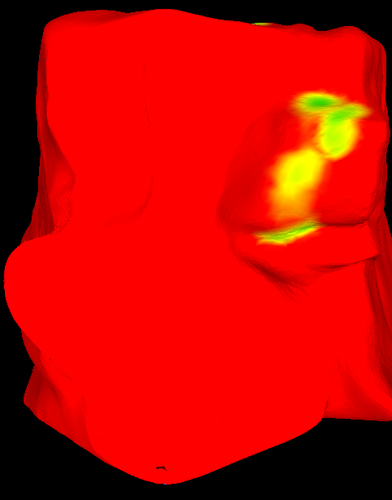}&
    \includegraphics[width=\mywidthx]{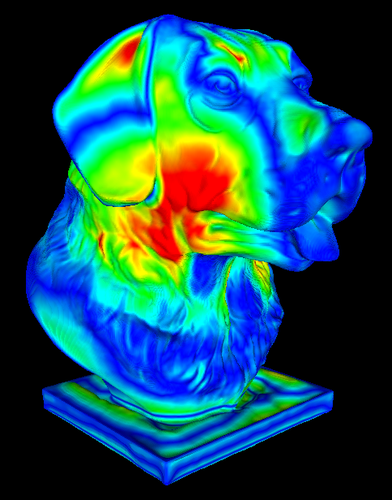}&
    \includegraphics[width=\mywidthx]{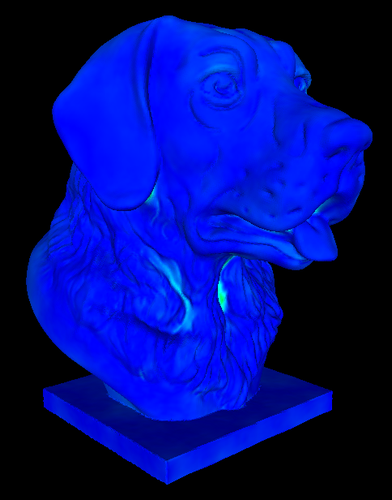}\\
    \rotatebox{90}{\dogB{}}&
    \includegraphics[width=\mywidthx]{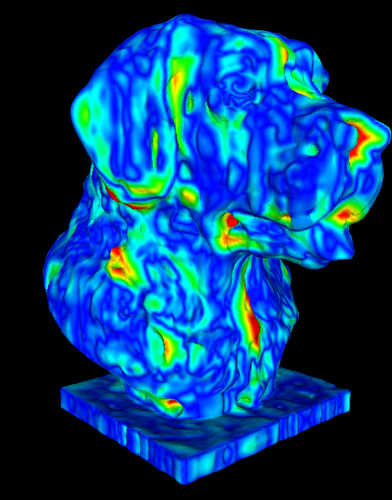} &
    \includegraphics[width=\mywidthx]{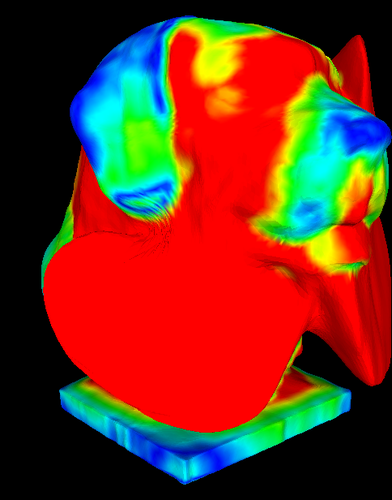}&  
    \includegraphics[width=\mywidthx]{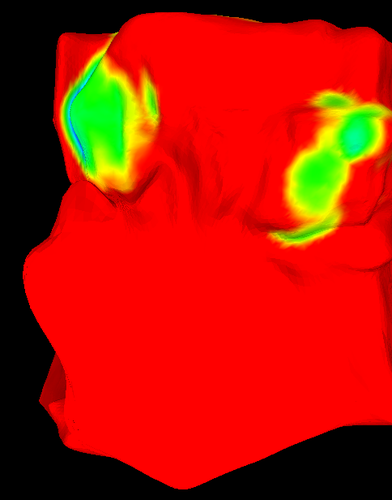}&
    \includegraphics[width=\mywidthx]{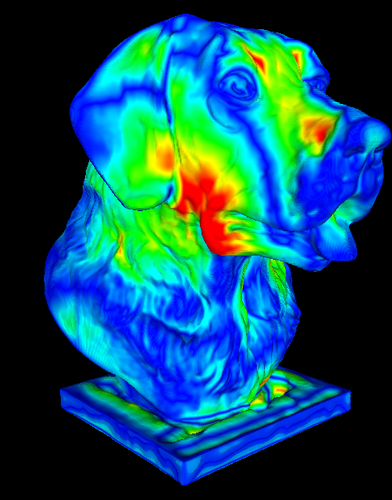}&
    \includegraphics[width=\mywidthx]{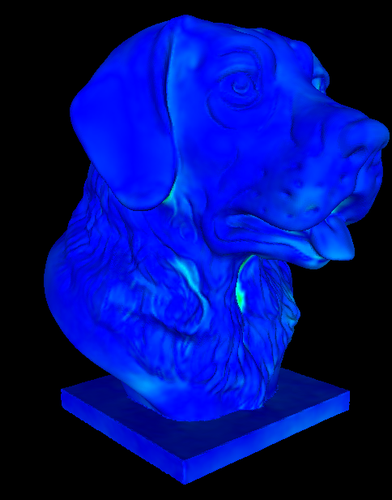}\\
    & \cite{Wu_2023_ICCV} & \cite{cheng2022diffeomorphic} (directional light) & \cite{cheng2022diffeomorphic} (point-light) & \cite{yang2022ps} & Ours
  \end{tabular} &
  \begin{tabular}{c}
    \colorbar[custom]{0}{0.05}{2}
  \end{tabular}
\end{tabular}
  \xcaption{Vertex-to-mesh distance error maps. Errors are truncated for better visibility.}
\label{fig:distance_error_maps}
\end{figure*}

\begin{figure*}[t]
  \small
  \newcommand{\mywidthc}{0.02\textwidth}
  \newcommand{\mywidthx}{0.164\textwidth}
  \newcolumntype{C}{ >{\centering\arraybackslash} m{\mywidthc} }
  \newcolumntype{X}{ >{\centering\arraybackslash} m{\mywidthx} }
  \newcommand{\tabelt}[1]{\hfil\hbox to 0pt{\hss #1 \hss}\hfil}
  \setlength\tabcolsep{1pt} %
\begin{tabular}{cc}
  \begin{tabular}{cXXXXX}
    \rotatebox{90}{\girlA{}}&
    \includegraphics[width=\mywidthx]{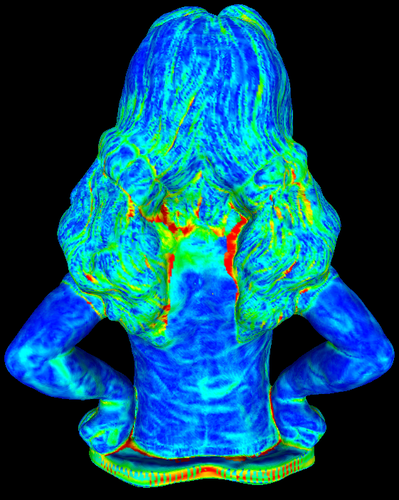} &
    \includegraphics[width=\mywidthx]{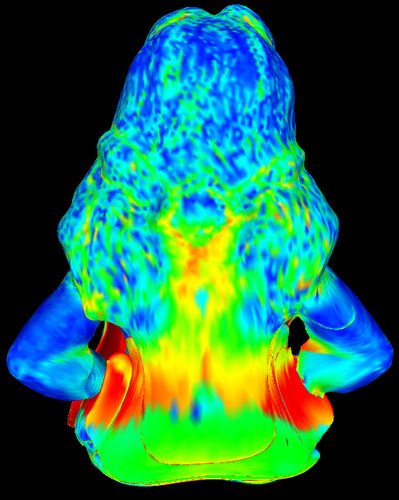}&  
    \includegraphics[width=\mywidthx]{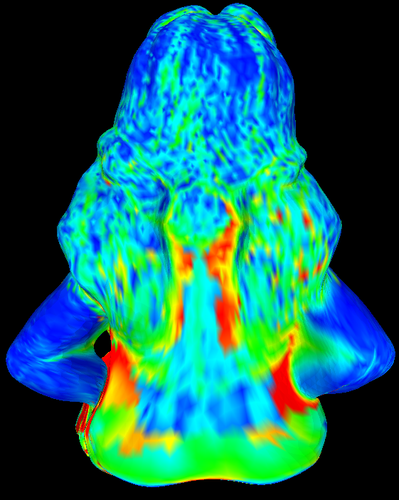}&
    \includegraphics[width=\mywidthx]{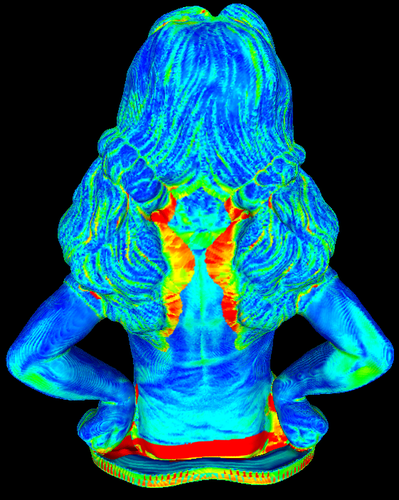}&
    \includegraphics[width=\mywidthx]{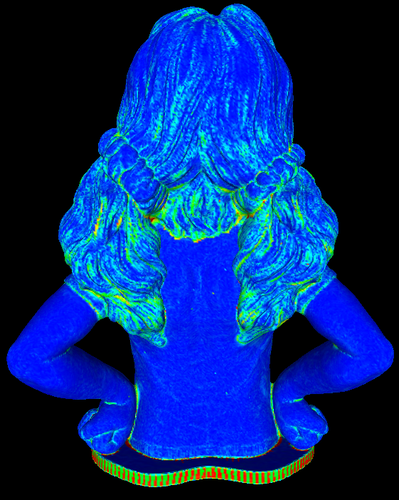}\\
    \rotatebox{90}{\girlB{}}&
    \includegraphics[width=\mywidthx]{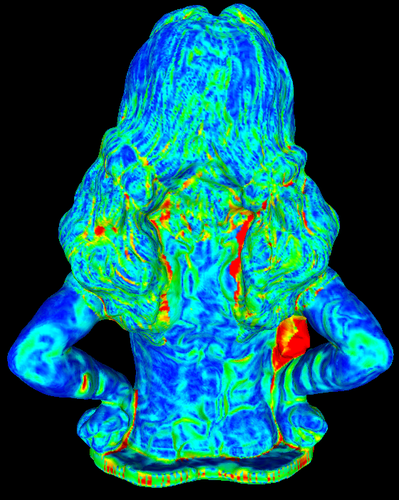} &
    \includegraphics[width=\mywidthx]{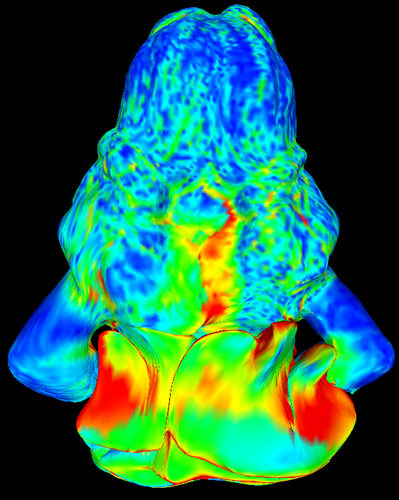}&  
    \includegraphics[width=\mywidthx]{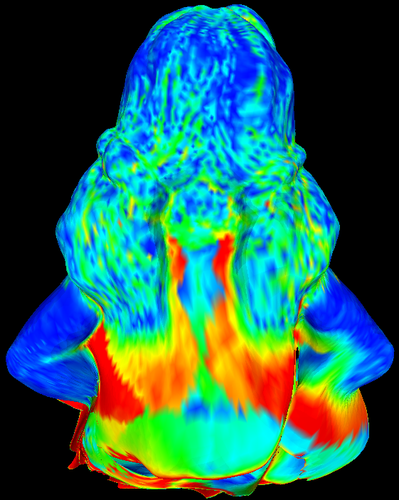}&
    \includegraphics[width=\mywidthx]{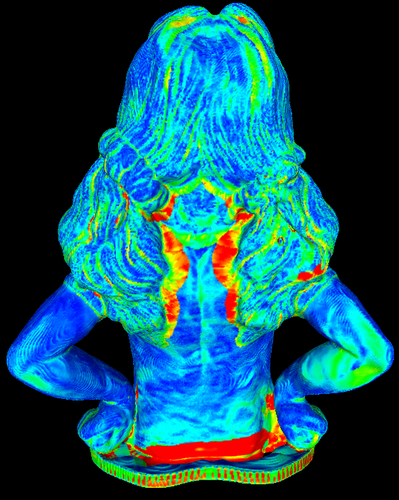}&
    \includegraphics[width=\mywidthx]{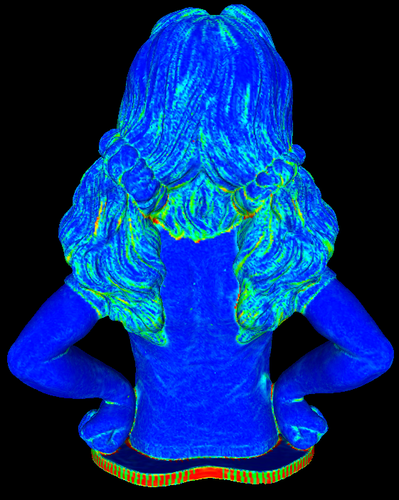}\\
    \rotatebox{90}{\dogA{}}&
    \includegraphics[width=\mywidthx]{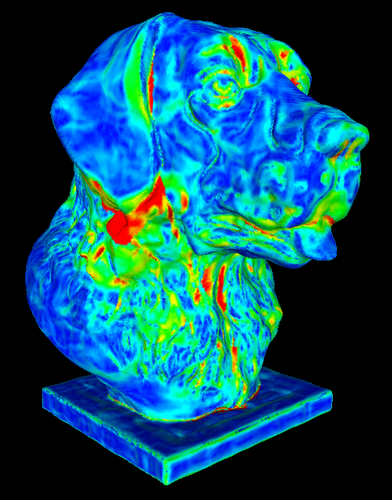} &
    \includegraphics[width=\mywidthx]{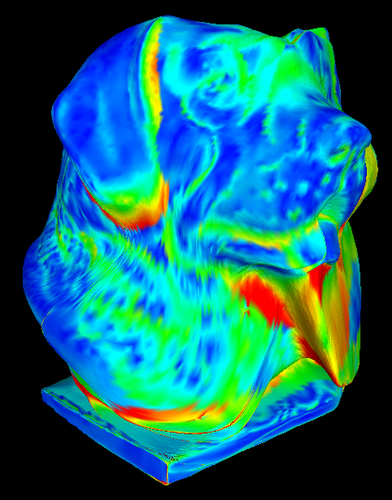}&  
    \includegraphics[width=\mywidthx]{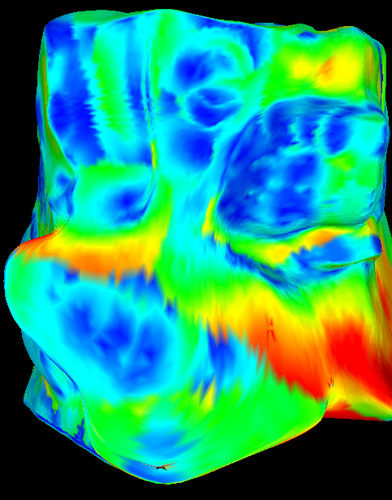}&
    \includegraphics[width=\mywidthx]{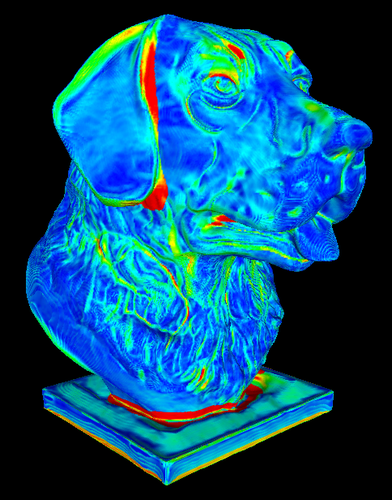}&
    \includegraphics[width=\mywidthx]{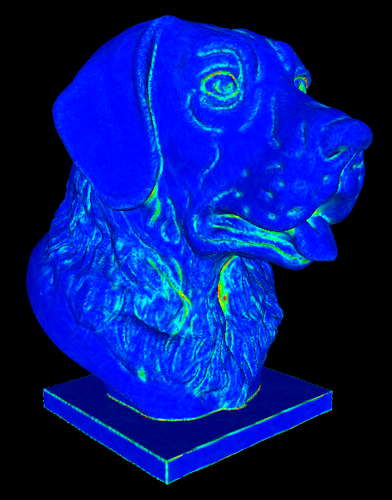}\\
    \rotatebox{90}{\dogB{}}&
    \includegraphics[width=\mywidthx]{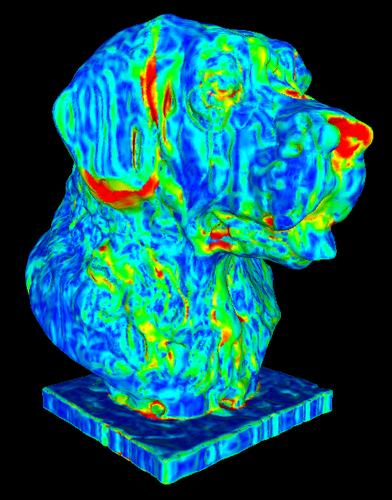} &
    \includegraphics[width=\mywidthx]{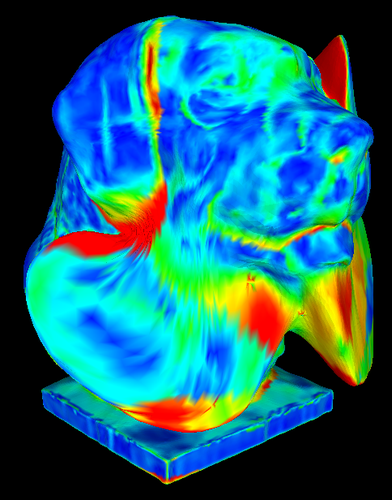}&  
    \includegraphics[width=\mywidthx]{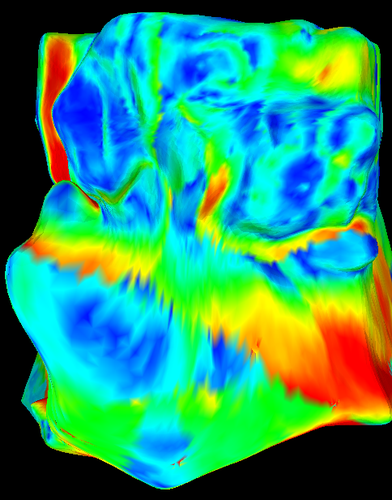}&
    \includegraphics[width=\mywidthx]{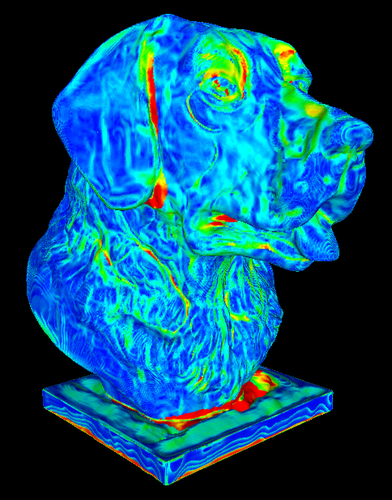}&
    \includegraphics[width=\mywidthx]{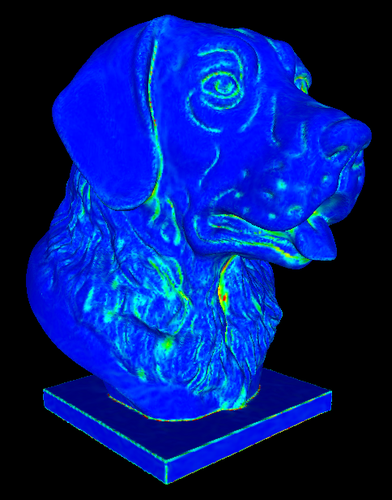}\\
    & \cite{Wu_2023_ICCV} & \cite{cheng2022diffeomorphic} (directional light) & \cite{cheng2022diffeomorphic} (point-light) & \cite{yang2022ps} & Ours
  \end{tabular} &
  \begin{tabular}{c}
    \colorbar[custom]{0}{90}{2}
  \end{tabular}
  \end{tabular}
  \xcaption{Angular error maps. Note that for \girlA{} and \girlB{}, a region of the plate at the bottom has significant errors for all approaches. This is due to the fact that in the ground truth mesh, vertices are only on the edges at that region, thus the angular error is not accurate there.}
\label{fig:angular_error_maps}
\end{figure*}

For a better appreciation of the quality of the full $3$D reconstructions shown in the main paper, we show both the vertex-to-mesh distance and angular error maps in \cref{fig:distance_error_maps} and \cref{fig:angular_error_maps} respectively. We can see that our approach performs much better than the baseline at both the coarse and fine levels. Hence, it not only produces visually more pleasant reconstructions as can be seen in the main paper, but also with much higher fidelity.

\vspace{-0.5mm}
\section{Additional results}
\begin{figure*}[t]
  \small
  \newcommand{\mywidthc}{0.02\textwidth}
  \newcommand{\mywidthx}{0.164\textwidth}
  \newcolumntype{C}{ >{\centering\arraybackslash} m{\mywidthc} }
  \newcolumntype{X}{ >{\centering\arraybackslash} m{\mywidthx} }
  \newcommand{\tabelt}[1]{\hfil\hbox to 0pt{\hss #1 \hss}\hfil}
  \setlength\tabcolsep{1pt} %
  \begin{tabular}{XXXXXX}
    \begin{tikzpicture}[spy using outlines={rectangle,connect spies}]
      \node[anchor=south west,inner sep=0]  at (0,0) {\includegraphics[width=\mywidthx]{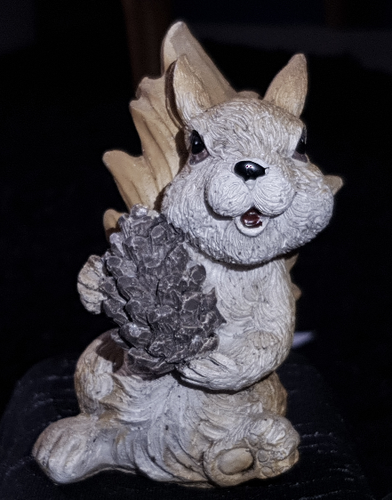}};
      \spy[color=green,width=0.9cm,height=0.95cm,magnification=1.7] on (1.3,0.95) in node [right] at (0.05,3.1);
    \end{tikzpicture}& 
    \begin{tikzpicture}[spy using outlines={rectangle,connect spies}]
      \node[anchor=south west,inner sep=0]  at (0,0) {\includegraphics[width=\mywidthx]{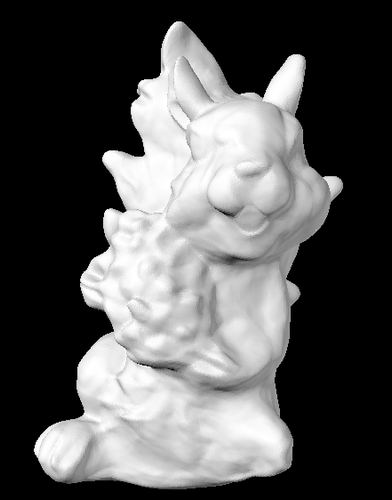}};
      \spy[color=green,width=0.9cm,height=0.95cm,magnification=1.7] on (1.3,0.95) in node [right] at (0.05,3.1);
    \end{tikzpicture}&  
    \begin{tikzpicture}[spy using outlines={rectangle,connect spies}]
      \node[anchor=south west,inner sep=0]  at (0,0) {\includegraphics[width=\mywidthx]{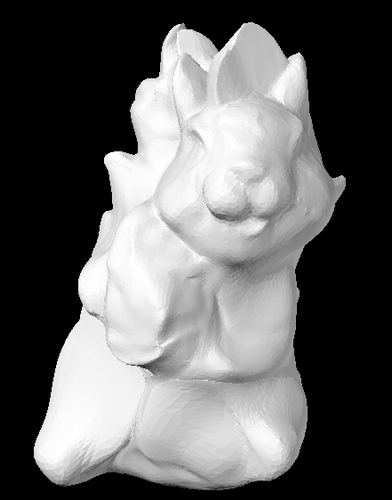}};
      \spy[color=green,width=0.9cm,height=0.95cm,magnification=1.7] on (1.3,0.95) in node [right] at (0.05,3.1);
    \end{tikzpicture}&  
    \begin{tikzpicture}[spy using outlines={rectangle,connect spies}]
      \node[anchor=south west,inner sep=0]  at (0,0) {\includegraphics[width=\mywidthx]{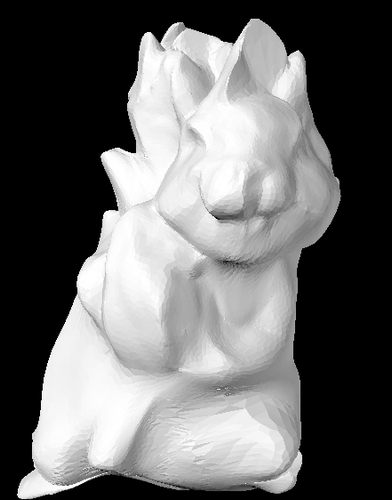}};
      \spy[color=green,width=0.9cm,height=0.95cm,magnification=1.7] on (1.3,0.95) in node [right] at (0.05,3.1);
    \end{tikzpicture}&
    \begin{tikzpicture}[spy using outlines={rectangle,connect spies}]
      \node[anchor=south west,inner sep=0]  at (0,0) {\includegraphics[width=\mywidthx]{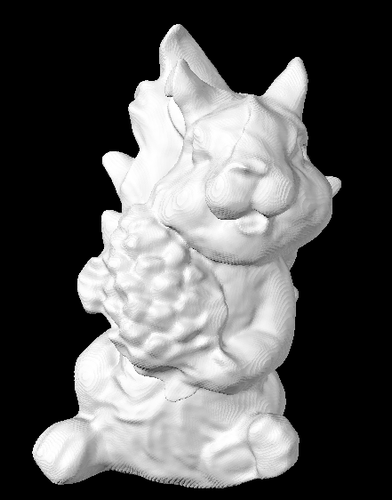}};
      \spy[color=green,width=0.9cm,height=0.95cm,magnification=1.7] on (1.3,0.95) in node [right] at (0.05,3.1);
    \end{tikzpicture}&
    \begin{tikzpicture}[spy using outlines={rectangle,connect spies}]
      \node[anchor=south west,inner sep=0]  at (0,0) {\includegraphics[width=\mywidthx]{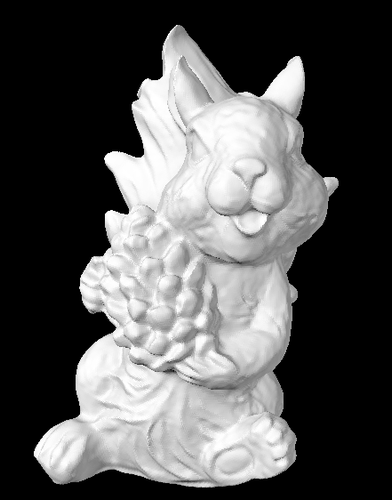}};
      \spy[color=green,width=0.9cm,height=0.95cm,magnification=1.7] on (1.3,0.95) in node [right] at (0.05,3.1);
    \end{tikzpicture}\\
    \begin{tikzpicture}[spy using outlines={rectangle,connect spies}]
      \node[anchor=south west,inner sep=0]  at (0,0) {\includegraphics[width=\mywidthx]{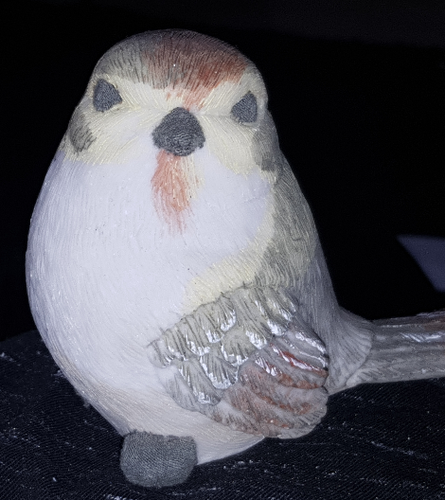}};
      \spy[color=green,width=0.9cm,height=0.9cm,magnification=1.7] on (1.3,1.0) in node [right] at (1.9,2.7);
    \end{tikzpicture}& 
    \begin{tikzpicture}[spy using outlines={rectangle,connect spies}]
      \node[anchor=south west,inner sep=0]  at (0,0) {\includegraphics[width=\mywidthx]{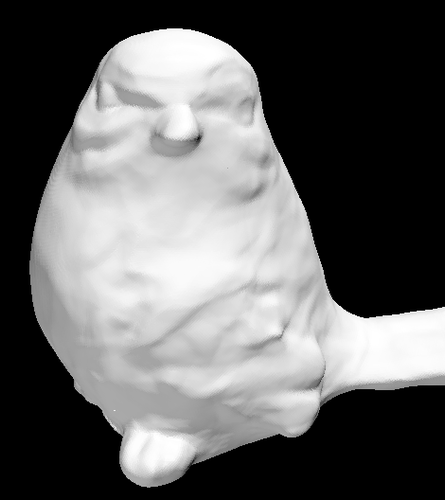}};
      \spy[color=green,width=0.9cm,height=0.9cm,magnification=1.7] on (1.3,1.0) in node [right] at (1.9,2.7);
    \end{tikzpicture}& 
    \begin{tikzpicture}[spy using outlines={rectangle,connect spies}]
      \node[anchor=south west,inner sep=0]  at (0,0) {\includegraphics[width=\mywidthx]{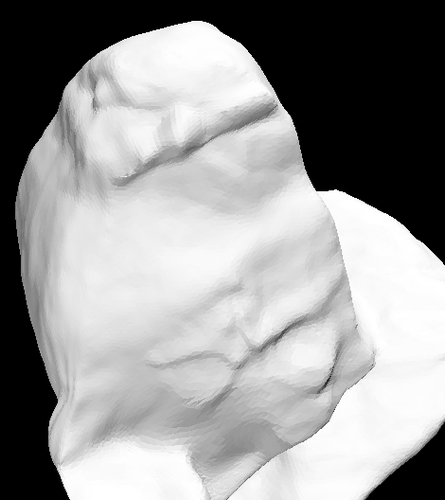}};
      \spy[color=green,width=0.9cm,height=0.9cm,magnification=1.7] on (1.3,1.0) in node [right] at (1.9,2.7);
    \end{tikzpicture}& 
    \begin{tikzpicture}[spy using outlines={rectangle,connect spies}]
      \node[anchor=south west,inner sep=0]  at (0,0) {\includegraphics[width=\mywidthx]{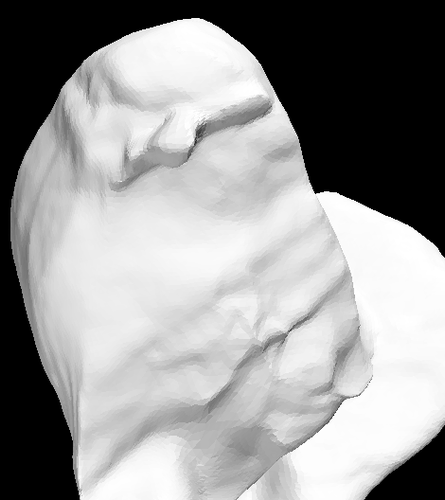}};
      \spy[color=green,width=0.9cm,height=0.9cm,magnification=1.7] on (1.3,1.0) in node [right] at (1.9,2.7);
    \end{tikzpicture}&
    \begin{tikzpicture}[spy using outlines={rectangle,connect spies}]
      \node[anchor=south west,inner sep=0]  at (0,0) {\includegraphics[width=\mywidthx]{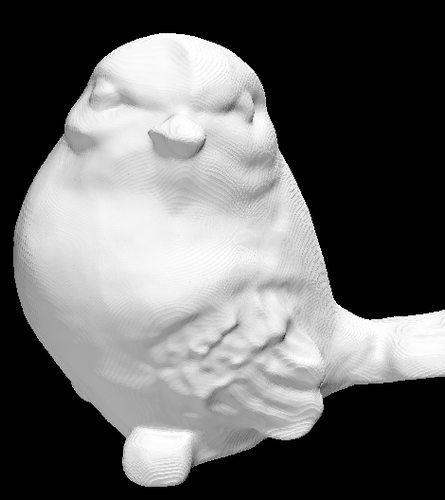} };
      \spy[color=green,width=0.9cm,height=0.9cm,magnification=1.7] on (1.3,1.0) in node [right] at (1.9,2.7);
    \end{tikzpicture}&
    \begin{tikzpicture}[spy using outlines={rectangle,connect spies}]
      \node[anchor=south west,inner sep=0]  at (0,0) {\includegraphics[width=\mywidthx]{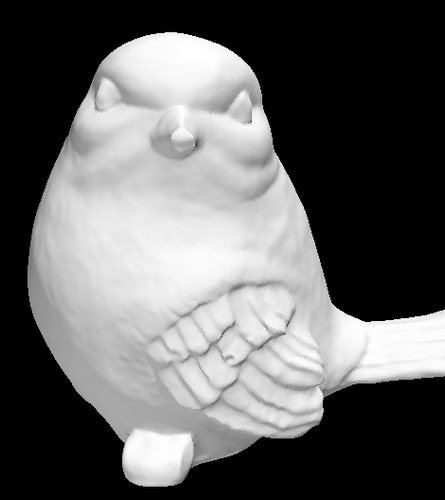}};
      \spy[color=green,width=0.9cm,height=0.9cm,magnification=1.7] on (1.3,1.0) in node [right] at (1.9,2.7);
    \end{tikzpicture}\\
	\begin{tikzpicture}[spy using outlines={rectangle,connect spies}]
      \node[anchor=south west,inner sep=0]  at (0,0) {\includegraphics[width=\mywidthx]{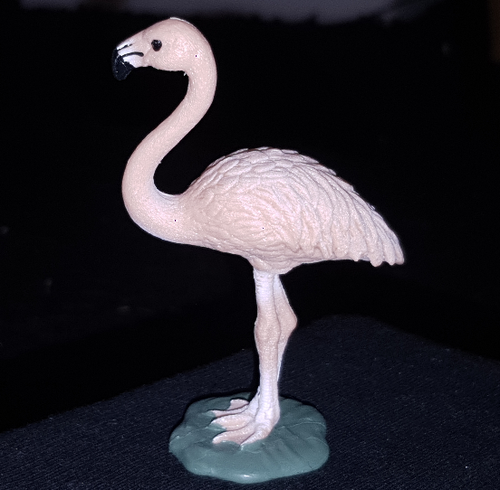}};
      \spy[color=green,width=0.9cm,height=0.6cm,magnification=1.7] on (1.45,1.55) in node [right] at (1.95,2.2);
    \end{tikzpicture}& 
    \begin{tikzpicture}[spy using outlines={rectangle,connect spies}]
      \node[anchor=south west,inner sep=0]  at (0,0) {\includegraphics[width=\mywidthx]{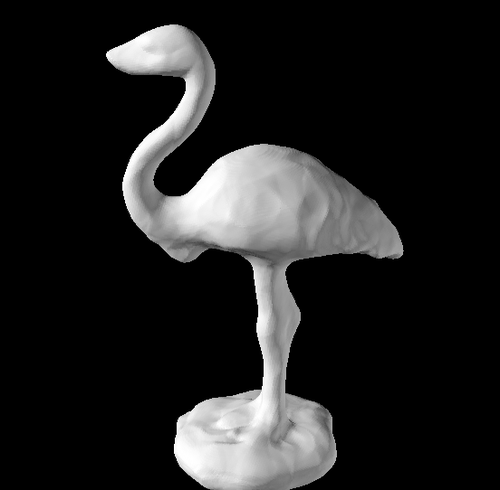}};
      \spy[color=green,width=0.9cm,height=0.6cm,magnification=1.7] on (1.45,1.55) in node [right] at (1.95,2.2);
    \end{tikzpicture}&  
    \begin{tikzpicture}[spy using outlines={rectangle,connect spies}]
      \node[anchor=south west,inner sep=0]  at (0,0) {\includegraphics[width=\mywidthx]{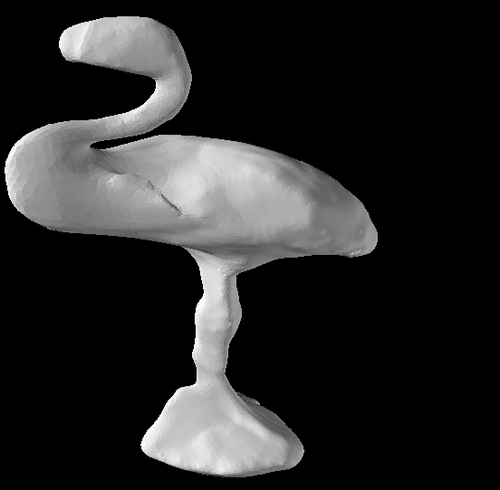}};
      \spy[color=green,width=0.9cm,height=0.6cm,magnification=1.7] on (1.45,1.55) in node [right] at (1.95,2.2);
    \end{tikzpicture}&  
    \begin{tikzpicture}[spy using outlines={rectangle,connect spies}]
      \node[anchor=south west,inner sep=0]  at (0,0) {\includegraphics[width=\mywidthx]{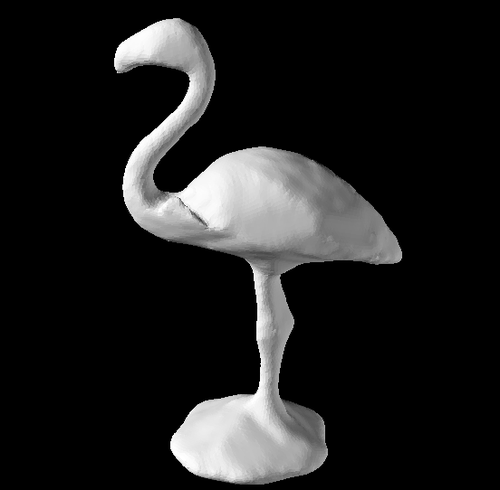}};
      \spy[color=green,width=0.9cm,height=0.6cm,magnification=1.7] on (1.45,1.55) in node [right] at (1.95,2.2);
    \end{tikzpicture}&
    \begin{tikzpicture}[spy using outlines={rectangle,connect spies}]
      \node[anchor=south west,inner sep=0]  at (0,0) {\includegraphics[width=\mywidthx]{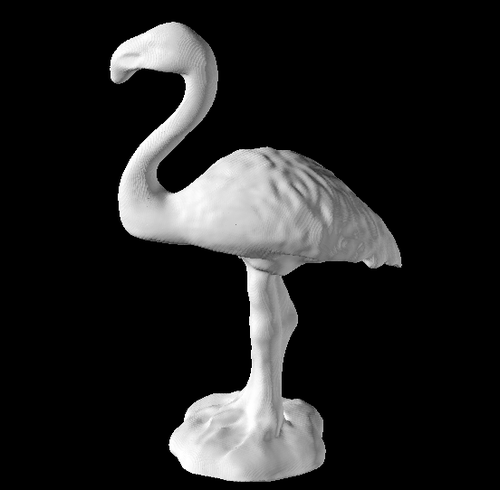}};
      \spy[color=green,width=0.9cm,height=0.6cm,magnification=1.7] on (1.45,1.55) in node [right] at (1.95,2.2);
    \end{tikzpicture}&
    \begin{tikzpicture}[spy using outlines={rectangle,connect spies}]
      \node[anchor=south west,inner sep=0]  at (0,0) {\includegraphics[width=\mywidthx]{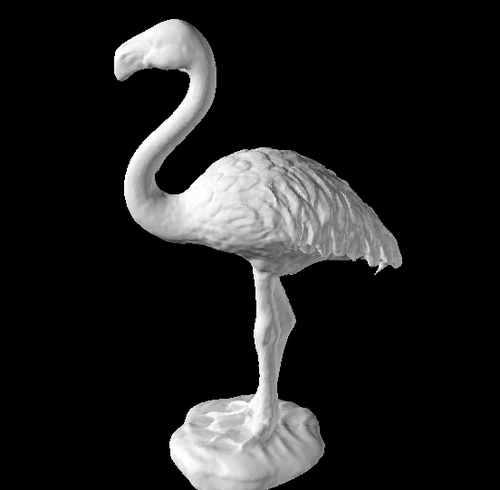}};
      \spy[color=green,width=0.9cm,height=0.6cm,magnification=1.7] on (1.45,1.55) in node [right] at (1.95,2.2);
    \end{tikzpicture}\\
	\includegraphics[width=\mywidthx]{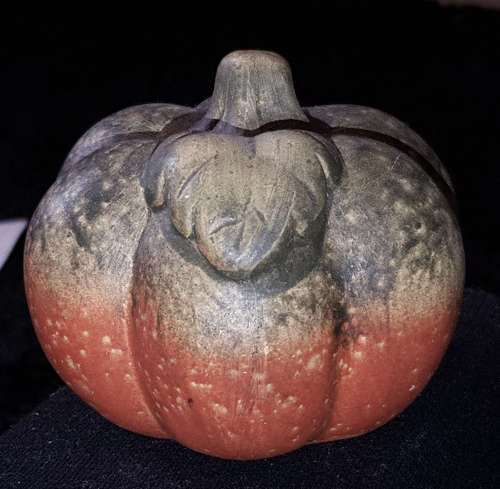}& 
    \includegraphics[width=\mywidthx]{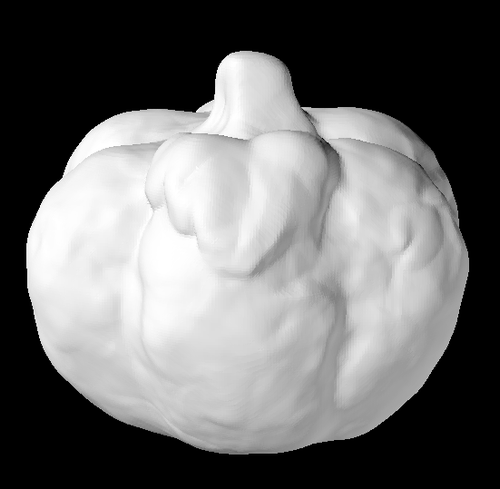}&  
    \includegraphics[width=\mywidthx]{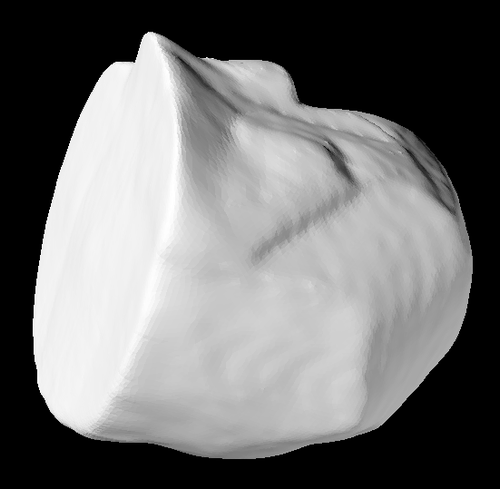}&  
    \includegraphics[width=\mywidthx]{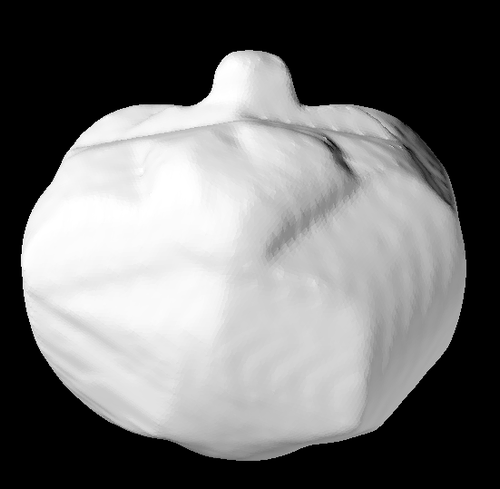}&
    \includegraphics[width=\mywidthx]{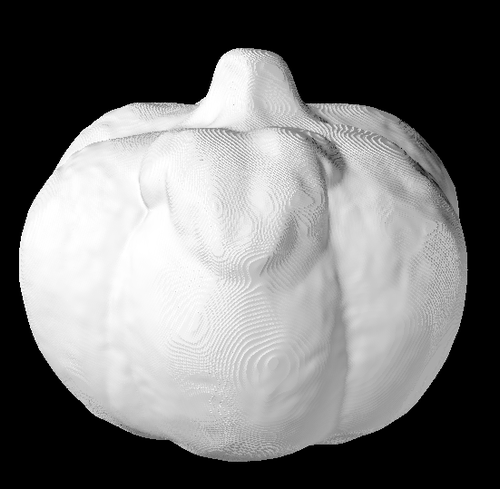}&
    \includegraphics[width=\mywidthx]{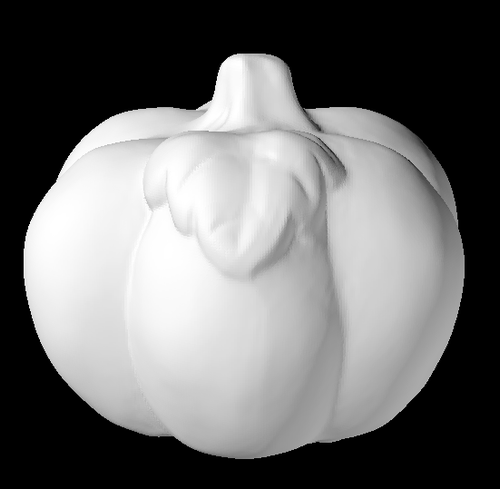}\\
    Input image & \cite{Wu_2023_ICCV} & \cite{cheng2022diffeomorphic} (directional light) & \cite{cheng2022diffeomorphic} (point-light) & \cite{yang2022ps} & Ours
  \end{tabular}
  \xcaption{Full 3D reconstruction of real objects from 6 viewpoints.}
\label{fig:complementary_results}
\end{figure*}

\begin{figure*}[t]
  \small
  \centering
  \newcommand{\mywidthc}{0.02\textwidth}
  \newcommand{\mywidthx}{0.164\textwidth}
  \newcolumntype{C}{ >{\centering\arraybackslash} m{\mywidthc} }
  \newcolumntype{X}{ >{\centering\arraybackslash} m{\mywidthx} }
  \newcommand{\tabelt}[1]{\hfil\hbox to 0pt{\hss #1 \hss}\hfil}
  \setlength\tabcolsep{1pt} %
  \begin{tabular}{XXXX}
	\includegraphics[width=\mywidthx]{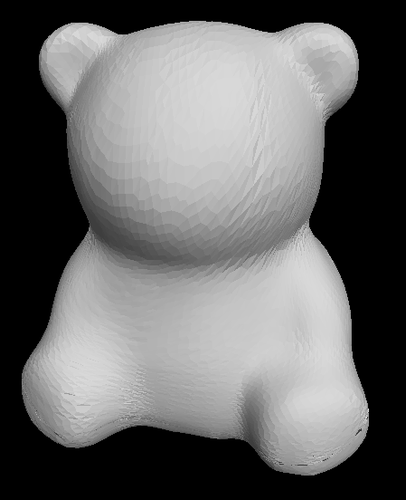}& 
    \includegraphics[width=\mywidthx]{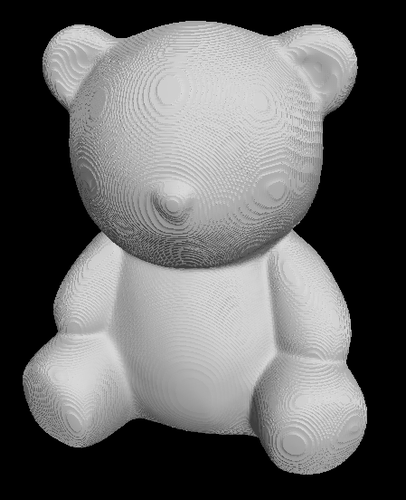}&  
    \includegraphics[width=\mywidthx]{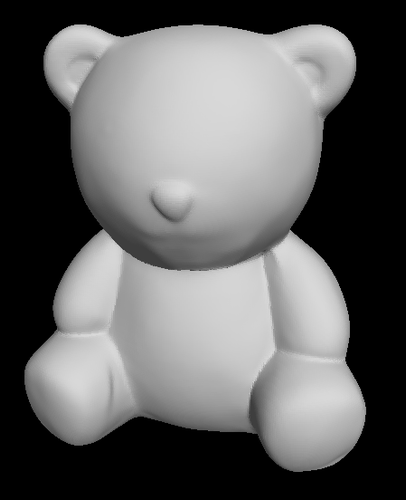}&
    \includegraphics[width=\mywidthx]{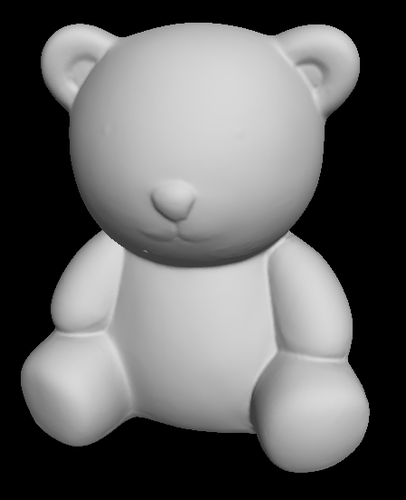}\\
	\includegraphics[width=\mywidthx]{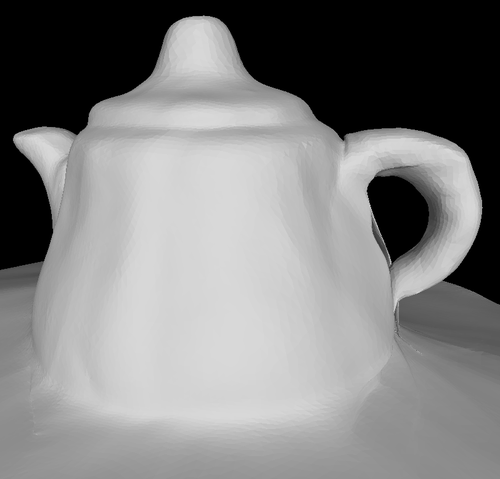}& 
    \includegraphics[width=\mywidthx]{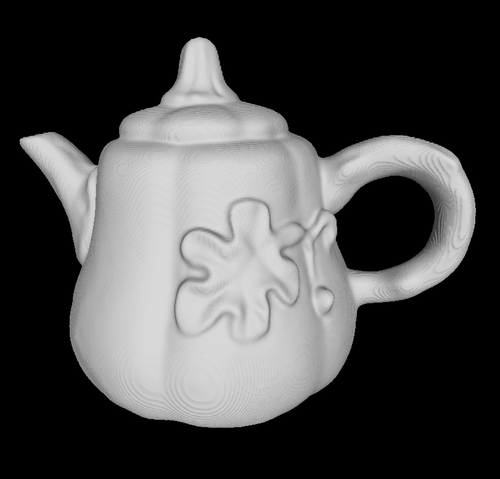}&  
    \includegraphics[width=\mywidthx]{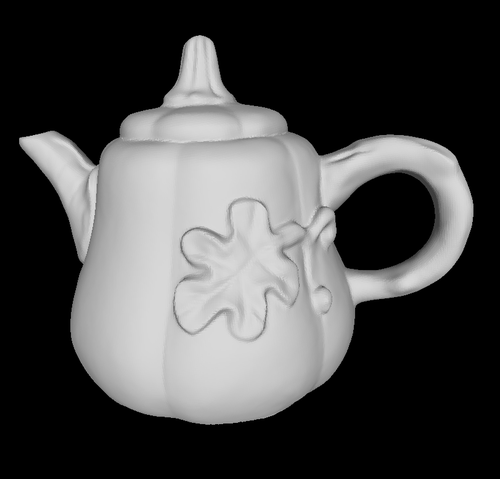}&
    \includegraphics[width=\mywidthx]{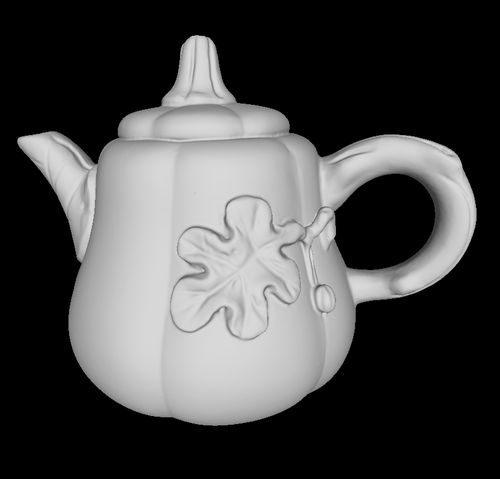}\\
	\includegraphics[width=\mywidthx]{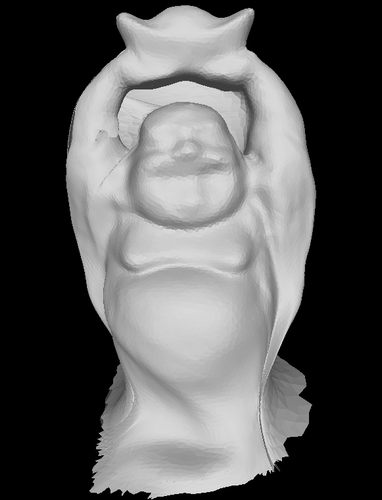}& 
    \includegraphics[width=\mywidthx]{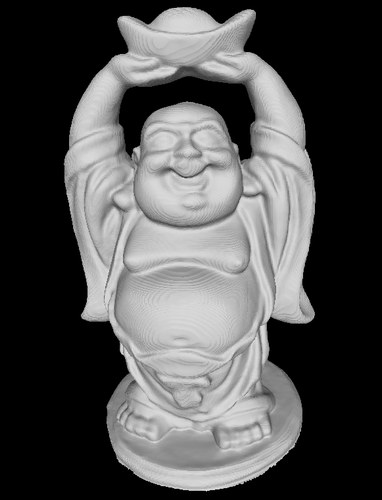}&  
    \includegraphics[width=\mywidthx]{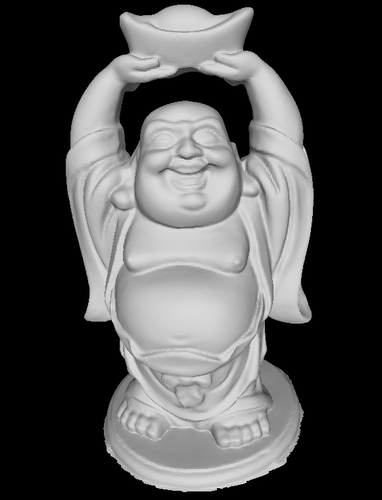}&
    \includegraphics[width=\mywidthx]{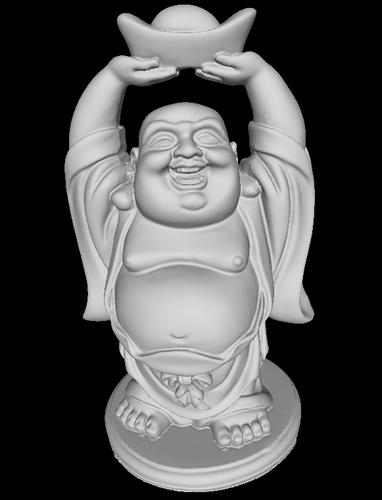}\\
	\includegraphics[width=\mywidthx]{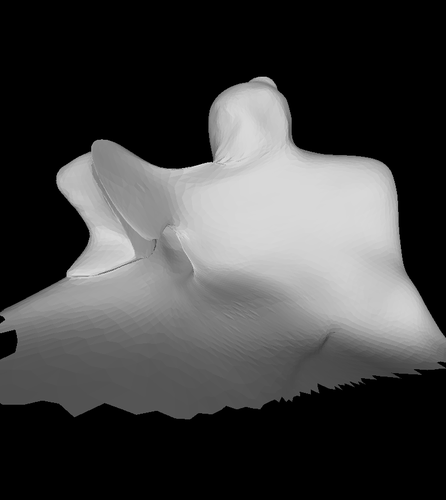}& 
    \includegraphics[width=\mywidthx]{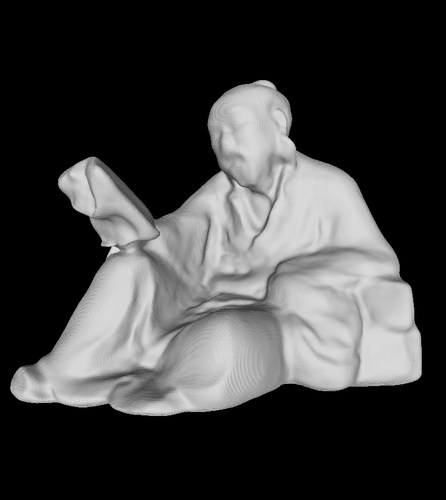}&  
    \includegraphics[width=\mywidthx]{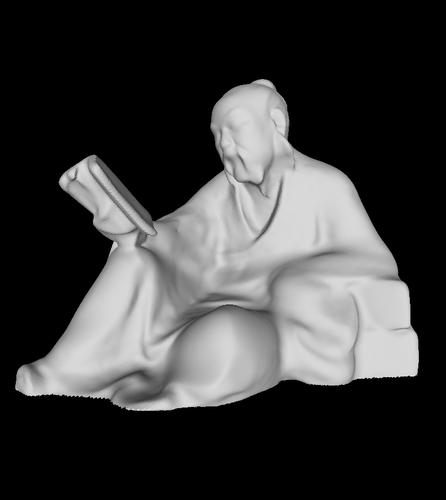}&
    \includegraphics[width=\mywidthx]{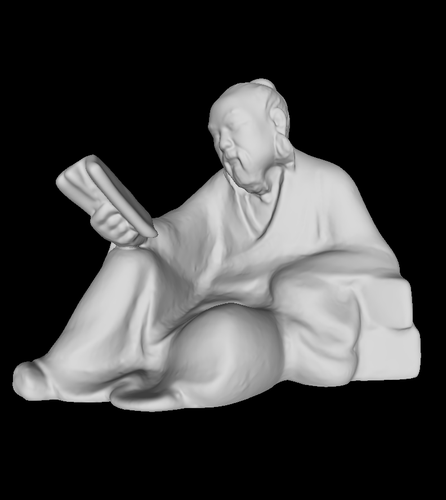}\\
	\includegraphics[width=\mywidthx]{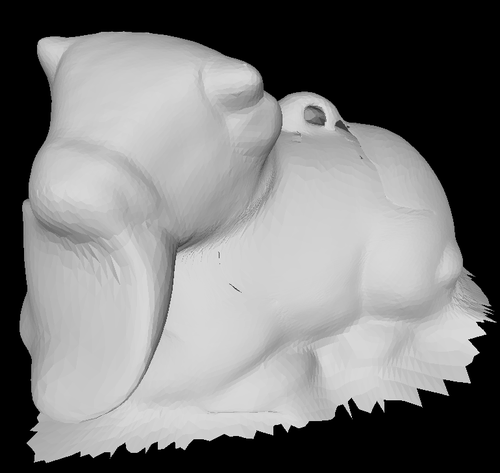}& 
    \includegraphics[width=\mywidthx]{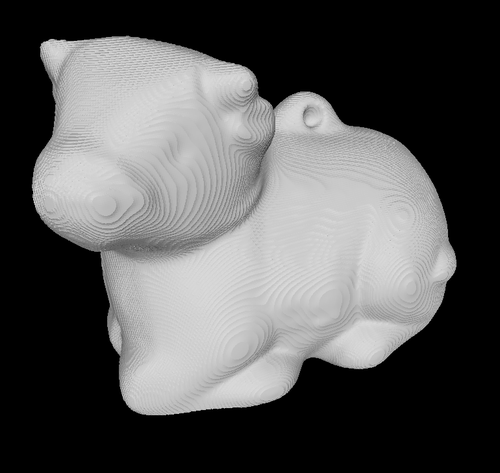}&  
    \includegraphics[width=\mywidthx]{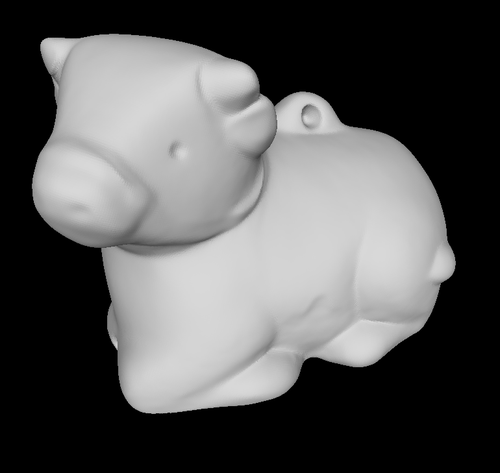}&
    \includegraphics[width=\mywidthx]{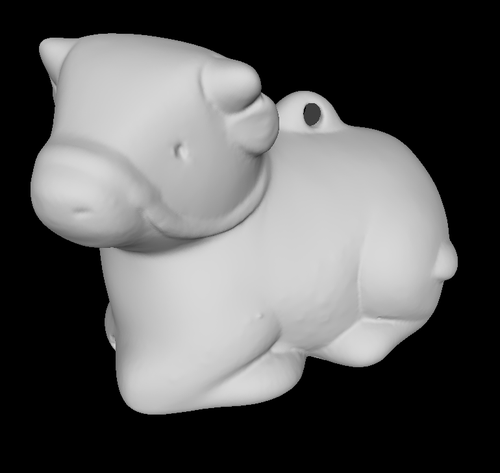}\\
    \cite{cheng2022diffeomorphic} & \cite{yang2022ps} & Ours & Ground truth
  \end{tabular}
  \xcaption{Results on the DiLiGenT-MV dataset \cite{li2020multi} using 6 viewpoints}
\label{fig:diligent_results}
\end{figure*}

We can see in \cref{fig:complementary_results} the reconstruction results of our real-world scans that were not shown in the main paper.
In order to further assess the quality of our framework on diverse materials, we performed an evaluation on the DiLiGenT-MV dataset \cite{li2020multi}. Despite being captured with distant light sources, thereby satisfying the directional lighting assumption used in the baseline, our framework still achieves the best results both quantitatively and qualitatively as can be seen respectively in \cref{tab:diligent} and \cref{fig:diligent_results}.
Finally, we also show some relighting results in \cref{fig:relighting_results}, together with the optimal diffuse albedo. This shows the validity of the estimated material parameters which can be successfully used for relighting, and indicates a proper disentanglement of the scene in terms of shape and material.

\begin{table}[t]
\centering
\setlength{\tabcolsep}{2pt}
\resizebox{0.8\linewidth}{!}{
\begin{tabular}{l|ccc|ccc}
~ & \multicolumn{3}{c|}{$\downarrow$MAE} & \multicolumn{3}{c}{$\downarrow$RMSE$\times 100$} \\
~ & \cite{cheng2022diffeomorphic} & \cite{yang2022ps} & ours & \cite{cheng2022diffeomorphic} & \cite{yang2022ps} & ours \\
\cline{1-7}
& &\\[-3.0mm]
\bear & 19.1 & 8.8 & \textbf{3.5} & 2.2 & 1.0 & \textbf{0.6} \\
\buddha & 39.6 & 13.9 & \textbf{10.8} & 2.5 & 0.7 & \textbf{0.5}\\
\pot & 29.2 & 9.2 & \textbf{5.1} & 10.2 & 0.7 & \textbf{0.5}\\
\reading & 34.7 & 11.9 & \textbf{7.1} & 10.4 & 1.1 & \textbf{0.9}\\
\cow & 25.3 & 8.9 & \textbf{3.6} & 6.1 & 0.9 & \textbf{0.5}
\end{tabular}
}
\xcaption{MAE and RMSE for the DiLiGenT-MV dataset \cite{li2020multi}.}
\label{tab:diligent}
\end{table}
\vspace{-0.5mm}
\section{Effect of the ratio of point light}
We further analyze the effect of the ratio of point light intensity on the quality of the result. This allows us to know how much ambient light can be handled by our approach while still providing accurate reconstructions. We remind that the total radiance can be decomposed into the sum of the point light radiance and the ambient light radiance, and we obtain the point light images by subtracting the input images with the ambient image. As discussed in section (3.2) of the main paper, one key issue with this strategy is that decreasing the amount of point light intensity yields point light images with a worse signal-to-noise ratio, which will inevitably affect the quality of the result. Consequently, we evaluate our approach on \dogB{} using the same five viewpoints as in the main paper to obtain a full $3$D reconstruction, with a point light intensity ratio ranging from $10\%$ to $100\%$ (dark room). We also consider two levels of noise, with standard deviations $\sigma \in \lbrace 0.02, 0.04 \rbrace$. As shown in \cref{tab:ablation_ratio} and \cref{fig:ablation_ratio}, the quality of the result indeed improves as expected when increasing the amount of point light intensity. Moreover, for a given desired accuracy, a higher amount of point light is required for a noisier sensor, since this last yields the worst signal-to-noise ratio for the point light images. Nevertheless, even with a significant amount of noise, a reasonable result can be obtained starting from $40\%$ of point light intensity, and a high accuracy with $70\%$ and above. As mentioned in section (3.2) of the main paper, satisfying those requirements in practice is highly facilitated by the fact that near point lights are handled properly by our framework, in contrast to the majority of photometric stereo frameworks which require distant lights.

\begin{table*}[t]
\footnotesize
\setlength{\tabcolsep}{2pt}
\resizebox{1.0\linewidth}{!}{
\begin{tabular}{l|cccccccccc|cccccccccc|}
~ & \multicolumn{10}{c|}{$\downarrow$RMSE$\times 1000$} & \multicolumn{10}{c|}{$\downarrow$MAE} \\
~ & $10\%$ & $20\%$ & $30\%$ & $40\%$ & $50\%$ & $60\%$ & $70\%$ & $80\%$ & $90\%$ & $100\%$ & $10\%$ & $20\%$ & $30\%$ & $40\%$ & $50\%$ & $60\%$ & $70\%$ & $80\%$ & $90\%$ & $100\%$\\
\cline{1-21}
& & & & & & & & & & & & & & & & & & & &\\[-2mm]
$\sigma=0.02$ & 10.3 & 5.6 & 4.3 & 3.9 & 3.3 & 3.4 & 3.2 & 3.2 & 2.9 & \textbf{2.6} & 12.7 & 7.8 & 6.3 & 5.7 & 5.2 & 5.0 & 4.7 & 4.8 & 4.6 & \textbf{4.4}\\
$\sigma=0.04$ & 16.1 & 10.3 & 6.5 & 5.2 & 4.5 & 3.9 & 3.6 & 3.6 & 3.5 & \textbf{3.2} & 16.7 & 12.8 & 9.4 & 7.8 & 6.9 & 6.2 & 5.8 & 5.5 & 5.4 & \textbf{5.0}
\end{tabular}
}
\xcaption{RMSE and MAE for different ratios of point light intensity, and two different levels of noise. RMSE is computed based on the vertex-to-mesh distance, and the MAE is computed using the angular error between the normals of a vertex and its closest point in the ground truth mesh.}
\label{tab:ablation_ratio}
\end{table*}
\begin{figure*}[t]
  \small
  \newcommand{\mywidthc}{0.02\textwidth}
  \newcommand{\mywidthx}{0.164\textwidth}
  \newcolumntype{C}{ >{\centering\arraybackslash} m{\mywidthc} }
  \newcolumntype{X}{ >{\centering\arraybackslash} m{\mywidthx} }
  \newcommand{\tabelt}[1]{\hfil\hbox to 0pt{\hss #1 \hss}\hfil}
  \setlength\tabcolsep{1pt} %
\begin{tabular}{cc}
  \begin{tabular}{CXXXX}
    \rotatebox{90}{$\sigma = 0.02$}&
    \includegraphics[width=\mywidthx]{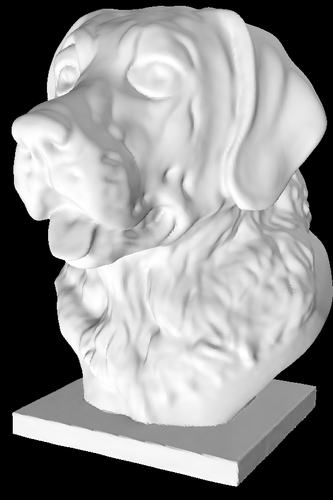} &
    \includegraphics[width=\mywidthx]{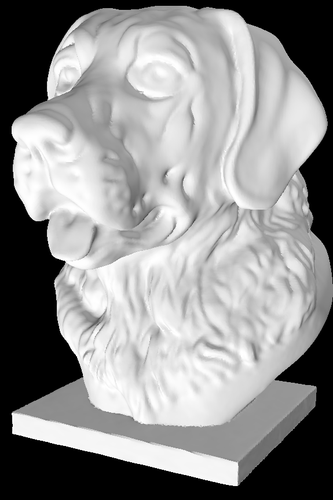}&  
    \includegraphics[width=\mywidthx]{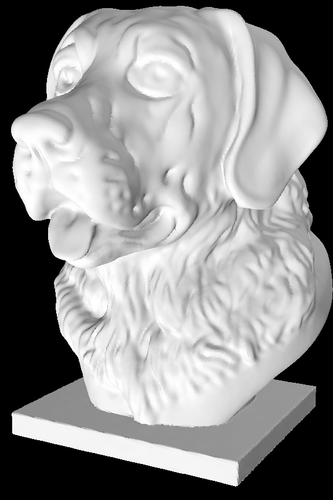}&
    \includegraphics[width=\mywidthx]{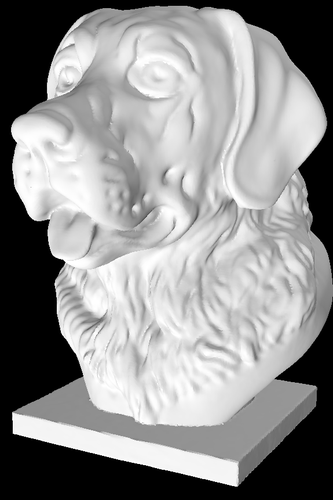}\\
    \rotatebox{90}{$\sigma = 0.04$}&
    \includegraphics[width=\mywidthx]{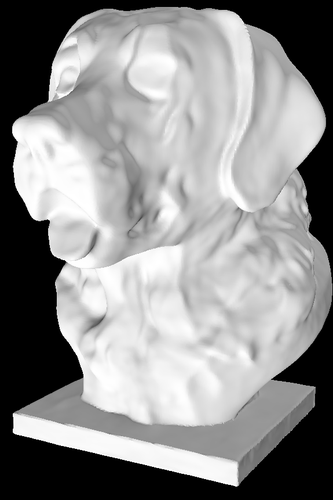} &
    \includegraphics[width=\mywidthx]{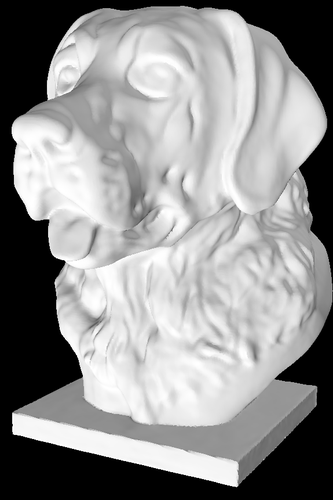}&  
    \includegraphics[width=\mywidthx]{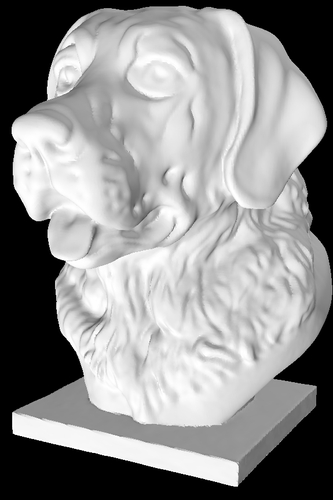}&
    \includegraphics[width=\mywidthx]{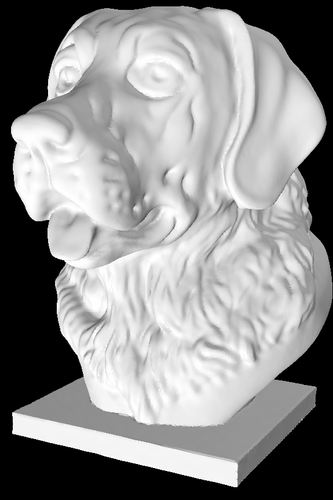}\\
    & $20\%$ & $40\%$ & $70\%$ & $100\%$ 
  \end{tabular} &
  \begin{tabular}{X}
  \includegraphics[width=\mywidthx]{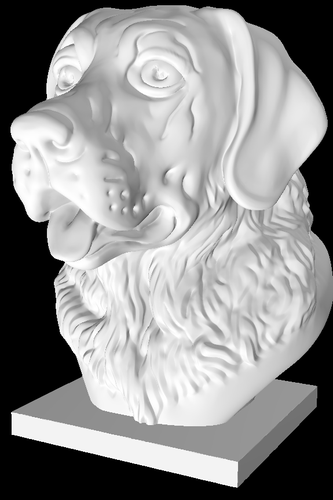} \\
  ground truth
  \end{tabular}
\end{tabular}
  \vspace{-4mm}
  \caption{Results on \dogB{} for different ratios of point light intensity. The first and second rows correspond to the results with Gaussian noise of standard deviation $\sigma = 0.02$ and $0.04$ respectively. }
\label{fig:ablation_ratio}
\vspace{4mm}
\begin{minipage}{.5\linewidth}
    \centering
	\small
  \newcommand{\mywidthcc}{0.02\textwidth}
  \newcommand{\mywidthxx}{0.28\textwidth}
  \newcolumntype{Y}{ >{\centering\arraybackslash} m{\mywidthxx} }
  \setlength\tabcolsep{1pt} %
  \begin{tabular}{YYY}
  \includegraphics[width=\mywidthxx]{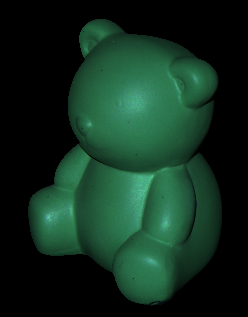} &
    \includegraphics[width=\mywidthxx]{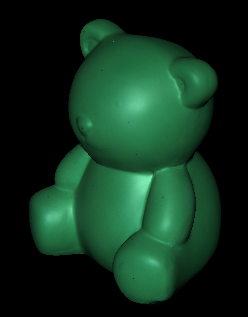} & 
	\includegraphics[width=\mywidthxx]{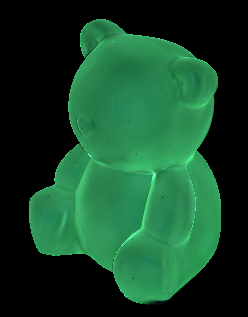}\\
	\includegraphics[width=\mywidthxx]{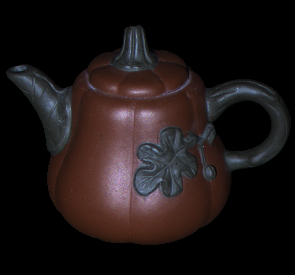} &
    \includegraphics[width=\mywidthxx]{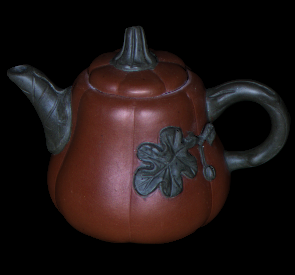} & 
	\includegraphics[width=\mywidthxx]{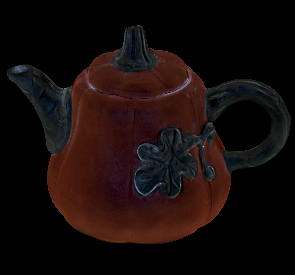}\\
	\includegraphics[width=\mywidthxx]{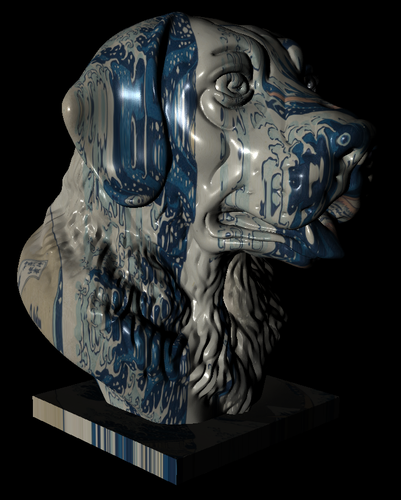} &
    \includegraphics[width=\mywidthxx]{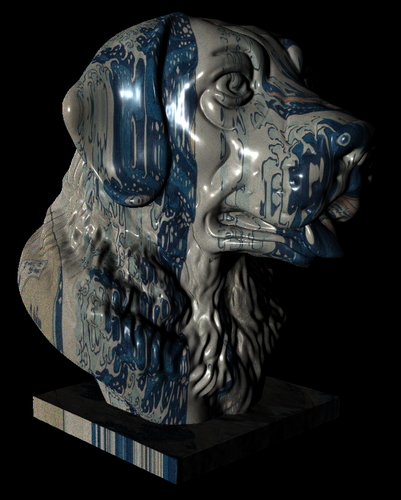} & 
	\includegraphics[width=\mywidthxx]{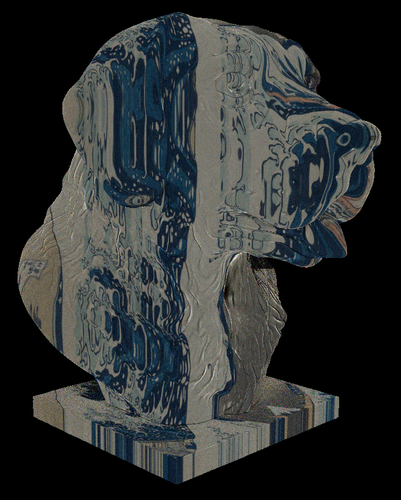}\\
    Ground truth & Our rendering & Optimal albedo
  \end{tabular}
  \vspace{-2mm}
  \caption{Relighting results.}
\label{fig:relighting_results}
  \end{minipage}
  \hspace{-5mm}
  \begin{minipage}{.5\linewidth}
    \centering
  \small
  \newcommand{\mywidthccc}{0.02\textwidth}
  \newcommand{\mywidthxxx}{0.155\textwidth}
  \setlength\tabcolsep{2pt} %
\begin{subfigure}{0.4\linewidth}
\begin{tabular}{l|ccc|}
~ & 5L & 10L & 20L\\
\cline{1-4}
& & & \\[-3.5mm]
4V & 7.7 & 6.1 & 5.5\\
6V & 6.4 & 5.0 & 4.6\\
8V & 6.0 & 4.6 & 4.3\\
12V & 5.1 & 4.2 & 3.9\\
\end{tabular}
\caption{\dogB{}}
\label{tab:dog_viewpoints_lights}
\end{subfigure}
\hspace{5mm}
\begin{subfigure}{0.4\linewidth}
\begin{tabular}{l|ccc|}
~ & 5L & 10L & 20L\\
\cline{1-4}
& & & \\[-3.5mm]
4V & 13.7 & 12.4 & 12.4\\
6V & 12.1 & 10.7 & 10.6\\
8V & 11.3 & 10.3 & 10.1\\
12V & 10.1 & 9.4 & 9.4\\
\end{tabular}
\caption{\girlB{}}
\label{tab:girl_viewpoints_lights}
\end{subfigure}
  \vspace{-2mm}
  \caption{MAE for different number of Viewpoints / Lights.}
\label{fig:table_ablation_viewpoints_lights}
  \end{minipage}
\end{figure*}

\section{Effect of the number of viewpoints and lights}
\cref{fig:table_ablation_viewpoints_lights} shows the MAE for both \dogB{} and \girlB{} using different numbers of viewpoints and light sources. Four viewpoints and five light sources allow to obtain a decent full $3$D reconstruction, and six viewpoints and ten light sources are already enough for a high quality result. 

\vspace{-0.5mm}
\section{Limitations}
The optimal diffuse albedo allows to obtain great $3$D reconstruction results in the most sparse scenarios. However, it is only defined for the viewpoints used for training, and is not multiview consistent, hindering novel view synthesis from arbitrary viewpoint.
On the other hand, this issue is mitigated with our ablation \OurAlbedoNet{} by using a neural diffuse albedo, at the cost of failing in some highly sparse scenarios. 
A straightforward solution would be to first use the optimal diffuse albedo strategy, then fix the geometry and specular parameters, and learn a neural diffuse albedo in a second stage. Successfully achieving multiview consistent diffuse albedo in the most sparse scenarios without relying on a second stage might increase the overall robustness, and is left as a future work.
Moreover, we presume the availability of camera poses, acknowledging the challenge of pose estimation, particularly in the context of sparse viewpoints.
A valuable extension of our work could be to address this assumption, \eg, based on~\cite{huang2023sc}.
Finally, our BRDF choice is limited to opaque, non-metallic objects. Expanding our framework beyond those materials represents an intriguing avenue for future exploration.
\clearpage

\end{document}